\documentclass{article}

\newif\ifsubmit
\submitfalse

\usepackage{iclr2018_conference, times}
\iclrfinalcopy
\usepackage{graphicx} % more modern
\usepackage{subcaption}
\usepackage{natbib}
\usepackage{algorithm}
\usepackage{algorithmic}
\usepackage{filecontents}
\usepackage{bm}
\usepackage{times}
\usepackage{booktabs}
\usepackage{graphicx}
\usepackage{amsmath}
\usepackage{amssymb}
\usepackage{url}
\usepackage{multirow}
\usepackage{color}
\newcommand{\argmax}{\mathop{\rm arg~max}\limits}

\usepackage{mathtools}

\usepackage{varwidth}
\usepackage{subcaption}
\usepackage{array}
\usepackage{hyperref}
\usepackage{empheq}
\usepackage{wrapfig}

\definecolor{medGray}{RGB}{230,230,230}

\renewcommand*{\bibfont}{\small}
\makeatletter
\newcommand{\figcaption}[1]{\def\@captype{figure}\caption{#1}}
\newcommand{\tblcaption}[1]{\def\@captype{table}\caption{#1}}
\makeatother

\ifsubmit
\newcommand\todo[1]{}
\newcommand\kwrite[1]{}
\newcommand\mwrite[1]{}
\newcommand\mnote[1]{}
\newcommand\ynote[1]{}
\newcommand\knote[1]{}
\else
\newcommand\todo[1]{\textcolor{red}{TODO: #1}}
\newcommand\mnote[1]{\textcolor{blue}{(MIYATO: #1)}}
\newcommand\mwrite[1]{\textcolor{green}{#1}}
\newcommand\kwrite[1]{\textcolor{brown}{#1}}
\newcommand\ynote[1]{\textcolor{blue}{(Yoshida: #1)}}
\newcommand\knote[1]{\textcolor{blue}{(Kataoka: #1)}}
\fi
\newcommand{\bbR}{\mathbb{R}}

\newcommand{\bmu}{{\bm u}}
\newcommand{\bmv}{{\bm v}}
\newcommand{\bmw}{{\bm w}}
\newcommand{\bmx}{{\bm x}}
\newcommand{\bmh}{{\bm h}}

\newcommand{\bmz}{{\bm z}}

\newcommand{\bmdelta}{{\bm \delta}}
\newcommand{\bmgamma}{{\bm \gamma}}
\newcommand{\bmmu}{{\bm \mu}}

\newcommand{\E}{\mathop{\mathrm{E}}}

\newenvironment{important}[2][]{
	\setkeys{EmphEqEnv}{#2}
	\setkeys{EmphEqOpt}{box={\setlength{\fboxsep}{10pt}\colorbox{medGray}},#1}
	\EmphEqMainEnv}
{\endEmphEqMainEnv}

\newenvironment{tight_itemize}{
\begin{itemize}
  \setlength{\itemsep}{0pt}
  \setlength{\parskip}{0pt}
  \setlength{\topsep}{0pt}
  \setlength{\partopsep}{0pt}
}{\end{itemize}}

\newenvironment{tight_enumerate}{
\begin{enumerate}
  \setlength{\itemsep}{0pt}
  \setlength{\parskip}{0pt}
  \setlength{\topsep}{0pt}
  \setlength{\partopsep}{0pt}
}{\end{enumerate}}

\title{Spectral Normalization\\ for Generative Adversarial Networks}

\author{Takeru Miyato\textsuperscript{1}, Toshiki Kataoka\textsuperscript{1}, Masanori Koyama\textsuperscript{2}, Yuichi Yoshida\textsuperscript{3}\\
\texttt{\{miyato, kataoka\}@preferred.jp} \\
\texttt{koyama.masanori@gmail.com} \\
\texttt{yyoshida@nii.ac.jp} \\
\textsuperscript{1}Preferred Networks, Inc.
\textsuperscript{2}Ritsumeikan University
\textsuperscript{3}National Institute of Informatics
}
\begin{document}
\maketitle
% \todo{TODO LIST}
% \begin{itemize}
% % % \item Explain conditional GANs
% % % \item Calc inception score and fid on STL-10
% % % \item 
% \end{itemize}
\begin{abstract}
One of the challenges in the study of generative adversarial networks is the instability of its training. 
In this paper, we propose a novel weight normalization technique called spectral normalization to stabilize the training of the discriminator.
Our new normalization technique is computationally light and easy to incorporate into existing implementations. 
We tested the efficacy of spectral normalization on CIFAR10, STL-10, and ILSVRC2012 dataset, and we experimentally confirmed that spectrally normalized GANs (SN-GANs) is capable of generating images of better or equal quality relative to the previous training stabilization techniques.  
The code with Chainer~\cite[]{tokui2015chainer}, generated images and pretrained models are available at \url{https://github.com/pfnet-research/sngan_projection}.
%To our knowledge, this is the first technique of its kind to successfully generate decent 1000-class images from ILSVRC2012 dataset using a single pair of a generator and a discriminator. 
\end{abstract}

\section{Introduction}
\textit{Generative adversarial networks}~(GANs)~\cite[]{goodfellow2014generative} have been enjoying considerable success as a framework of generative models in recent years, and it has been applied to numerous types of tasks and datasets~\cite[]{radford2015unsupervised, salimans2016improved, ho2016generative, li2017adversarial}.
In a nutshell, GANs are a framework to produce a model distribution that mimics a given target distribution,
and it consists of a generator that produces the model distribution and a discriminator that distinguishes the model distribution from the target.
The concept is to consecutively train the model distribution and the discriminator in turn,
with the goal of reducing the difference between the model distribution and the target distribution measured by the best discriminator possible at each step of the training.
GANs have been drawing attention in the machine learning community not only for its ability to learn highly structured probability distribution but also for its theoretically interesting aspects.
For example,~\cite[]{nowozin2016f, uehara2016generative, mohamed2016learning} revealed that the training of the discriminator amounts to the training of a good estimator for the density ratio between the model distribution and the target.
This is a perspective that opens the door to the methods of
\textit{implicit models}~\cite[]{mohamed2016learning, tran2017deep} that can be used to carry out variational optimization without the direct knowledge of the density function.

A persisting challenge in the training of GANs is the performance control of the discriminator. In high dimensional spaces, the density ratio estimation by the discriminator is often inaccurate and unstable during the training, and generator networks fail to learn the multimodal structure of the target distribution.
Even worse, when the support of the model distribution and the support of the target distribution are disjoint, there exists a discriminator that can perfectly distinguish the model distribution from the target~\cite[]{arjovsky2017towards}.
Once such discriminator is produced in this situation, the training of the generator comes to complete stop, because the derivative of the so-produced discriminator with respect to the input turns out to be 0.
This motivates us to introduce some form of restriction to the choice of the discriminator.

In this paper, we propose a novel weight normalization method called \textit{spectral normalization} that can stabilize the training of discriminator networks.
Our normalization enjoys following favorable properties.
\begin{tight_itemize}
\item Lipschitz constant is the only hyper-parameter to be tuned, and the algorithm does not require intensive tuning of the only hyper-parameter for satisfactory performance.
\item Implementation is simple and the additional computational cost is small.
\end{tight_itemize}
In fact, our normalization method also functioned well even without tuning Lipschitz constant, which is the only hyper parameter.
In this study, we provide explanations of the effectiveness of spectral normalization for GANs against other regularization techniques, such as weight normalization~\cite[]{salimans2016weight}, weight clipping~\cite[]{arjovsky2017wasserstein}, and gradient penalty~\cite[]{gulrajani2017improved}.
We also show that, in the absence of complimentary regularization techniques (e.g., batch normalization, weight decay and feature matching on the discriminator), spectral normalization can improve the sheer quality of the generated images better than weight normalization and gradient penalty.

\section{Method}
In this section, we will lay the theoretical groundwork for our proposed method.
Let us consider a simple discriminator made of a neural network of the following form, with the input $\bmx$:
\begin{align}
	f(\bmx, \theta) = W^{L+1} a_L (W^L (a_{L-1}(W^{L-1}(\dots a_1(W^1\bmx)\dots)))),
\end{align}
where $\theta:=\{W^1,\dots, W^L, W^{L+1} \}$ is the learning parameters set, $W^l\in \bbR^{d_l \times d_{l-1}}$, $W^{L+1}\in \bbR^{1\times{d_L}}$, and $a_l$ is an element-wise non-linear activation function. We omit the bias term of each layer for simplicity.
The final output of the discriminator is given by
\begin{align}
	D(\bmx, \theta) =  {\rm \mathcal{A}}(f(\bmx, \theta)),
\end{align}
where $\mathcal{A}$ is an activation function corresponding to the divergence of distance measure of the user's choice.
The standard formulation of GANs is given by
$$\min_{G} \max_{D} V(G, D)$$
where min and max of $G$ and $D$ are taken over the set of generator and discriminator functions, respectively.
The conventional form of $V(G,D)$~\cite[]{goodfellow2014generative} is given by
$\mathrm{E}_{\bmx\sim q_{\rm data}} [\log D(\bmx)] + \mathrm{E}_{\bmx'\sim p_G} [\log(1-D(\bmx'))], \label{eq:stdgan}$
where $q_{\rm data}$ is the data distribution and $p_G$ is the (model) generator distribution to be learned through the adversarial min-max optimization.
The activation function ${\rm \mathcal{A}}$ that is used in the $D$ of this expression is some continuous function with range $[0,1]$ (e.g, sigmoid function).
It is known that, for a fixed generator $G$, the optimal discriminator for this form of $V(G,D)$ is given by $D^*_G(\bmx) := q_{\rm data}(\bmx)/(q_{\rm data}(\bmx) + p_{G}(\bmx))$.

The machine learning community has been pointing out recently that the function space from which the discriminators are selected crucially affects the performance of GANs.
A number of works~\cite[]{uehara2016generative, qi2017loss, gulrajani2017improved} advocate the importance of Lipschitz continuity in assuring the boundedness of statistics.
For example, the optimal discriminator of GANs on the above standard formulation takes the form
\begin{align}
	D_{G}^*(\bmx) & = \frac{q_{\rm data}(\bmx)}{q_{\rm data}(\bmx)+p_G(\bmx)} = {\rm sigmoid}(f^*(\bmx)),
	\text{where }f^*(\bmx) = \log q_{\rm data}(\bmx) - \log p_G(\bmx),
\end{align}

and its derivative
\begin{align}
\nabla_\bmx f^*(x) = \frac{1}{q_{\rm data}(\bmx)}\nabla_{\bmx}q_{\rm data}(\bmx) - \frac{1}{p_G(\bmx)}\nabla_{\bmx}p_G(\bmx)
\end{align}
can be unbounded or even incomputable.
This prompts us to introduce some regularity condition to the derivative of $f(\bmx)$.

A particularly successful works in this array are~\cite[]{qi2017loss, arjovsky2017wasserstein, gulrajani2017improved}, which proposed methods to control the Lipschitz constant of the  discriminator by adding regularization terms defined on input examples $\bmx$.
We would follow their footsteps and search for the discriminator $D$ from the set of $K$-Lipschitz continuous functions, that is,
\begin{align}
	\argmax_{\|f\|_{\rm Lip}\leq K} V(G, D), \label{eq:lipschitz}
\end{align}
where we mean by $\|f\|_{\rm Lip}$ the smallest value $M$ such that $\|f(\bmx)-f(\bmx')\| / \|\bmx-\bmx'\|\leq M$ for any $\bmx,\bmx'$, with the norm being the $\ell_2$ norm.
% * <yoshida.yuuichi@gmail.com> 2017-10-13T05:59:22.007Z:
%
% > $\|f\|_{\rm Lip}$
%
% ^.
%\ynote{It's better to mention that $\|f\|_L = \|\nabla f\|_2$ (am I right?) or replace the constraint with $\|\nabla f(\bmx)\|_2 \leq K$ because we only discuss $\|\nabla f\|_2$ in the subsequent paragraph.}
%\knote{I agree with Yoshida-san. (To be precise, $\|f\|_{\rm Lip} = \sup_\bmx \|\nabla_\bmx f(\bmx)\|_2$.)
%Let's denote spectral norms only by $\sigma(W)$ (not by $\|W\|_2$, which emphasize that it denotes the spectral norm \emph{between L2 spaces}.)
%I'd like to mention the following at the beginning of \S{}2.1:
%it holds $\|f\|_{\rm Lip} = \sup_\bmx \sigma(\nabla_\bmx f(\bmx))$;
%therefore, for a linear layer $g(\bmh) = W\bmh + b$, we have $\|g\|_{\rm Lip} = \sup_\bmh \sigma((\nabla_\bmh g)(\bmh)) = \sup_\bmh \sigma(W) = \sigma(W) $.}\mnote{Thanks for your comments. I added the sentence for it.}
% DONE: \knote{$\|f\|_\mathrm{Lip}$}

While input based regularizations allow for relatively easy formulations based on samples, they also suffer from the fact that, they cannot impose regularization on the space outside of the supports of the generator and data distributions without introducing somewhat heuristic means.
A method we would introduce in this paper, called \textit{spectral normalization}, is a method that aims to skirt this issue by \textit{normalizing} the weight matrices using the technique devised by~\cite{yoshida2017spectral}.

\subsection{\label{subsec:SN}Spectral Normalization}
Our spectral normalization controls the Lipschitz constant of the discriminator function $f$ by literally constraining the spectral norm of each layer $g: \bmh_{in} \mapsto \bmh_{out}$. By definition, Lipschitz norm $\|g\|_{\rm Lip}$ is equal to $\sup_\bmh \sigma(\nabla g(\bmh))$, where $\sigma(A)$ is the spectral norm of the matrix $A$ ($L_2$ matrix norm of $A$)
\begin{align}
	\sigma(A) := \max_{\bmh: \bmh \neq {\bm 0}} \frac{\|A\bmh\|_2}{\|\bmh\|_2} = \max_{\|\bmh\|_2\leq 1} \|A\bmh\|_2, \label{eq:spnorm}
\end{align}
which is equivalent to the largest singular value of $A$.
Therefore, for a linear layer $g(\bmh) = W\bmh$, the norm is given by $\|g\|_{\rm Lip} = \sup_\bmh \sigma(\nabla g(\bmh)) = \sup_\bmh \sigma(W) = \sigma(W) $.
If the Lipschitz norm of the activation function $\|a_l\|_{\rm Lip}$ is equal to 1 \footnote{For examples, ReLU~\cite[]{jarrett2009best,nair2010rectified, glorot2011deep} and leaky ReLU~\cite[]{maas2013rectifier} satisfies the condition, and many popular activation functions satisfy $K$-Lipschitz constraint for some predefined $K$ as well.}, we can use the inequality $\|g_1\circ g_2\|_{\rm Lip}\leq\|g_1\|_{\rm Lip}\cdot\|g_2\|_{\rm Lip}$ to observe the following bound on $\|f\|_{\rm Lip}$:
\begin{align}
 	\|f\|_{\rm Lip} \leq & \|(\bmh_{L}\mapsto W^{L+1}\bmh_{L})\|_{\rm Lip}\cdot \|a_L\|_{\rm Lip}\cdot\|(\bmh_{L-1}\mapsto W^ L\bmh_{L-1})\|_{\rm Lip} \nonumber \\
    &\cdots \|a_1\|_{\rm Lip}\cdot\|(\bmh_0\mapsto W^ 1\bmh_0)\|_{\rm Lip} 
      = \prod_{l=1}^{L+1} \|(\bmh_{l-1}\mapsto W^l\bmh_{l-1})\|_{\rm Lip} = \prod_{l=1}^{L+1} \sigma(W^l). \label{eq:ineq_lip}
\end{align}
Our \textit{spectral normalization} normalizes the spectral norm of the weight matrix $W$ so that it satisfies the Lipschitz constraint $\sigma(W) = 1$:
\begin{align}
\bar{W}_{\rm SN}(W) := W / \sigma(W) \label{eq:sn}.
\end{align}
If we normalize each $W^l$ using~\eqref{eq:sn}, we can appeal to the inequality~\eqref{eq:ineq_lip} and the fact that $\sigma\left(\bar{W}_{\rm SN}(W)\right) = 1$ to see that $\|f\|_{\rm Lip}$ is bounded from above by $1$.

Here, we would like to emphasize the difference between our spectral normalization and spectral norm "regularization" introduced by ~\cite{yoshida2017spectral}.
Unlike our method, spectral norm "regularization" penalizes the spectral norm by adding explicit regularization term to the objective function. Their method is fundamentally different from our method in that they do not make an attempt to `set' the spectral norm to a designated value. Moreover, when we reorganize the derivative of our normalized cost function and rewrite our objective function \eqref{eq:grad_sn}, we see that our method is augmenting the cost function with a sample data \textit{dependent} regularization function. 
Spectral norm regularization, on the other hand, imposes sample data \textit{independent} regularization on the cost function,  just like L2 regularization and Lasso. 

\subsection{Fast Approximation of the Spectral Norm $\sigma(W)$}
As we mentioned above, the spectral norm $\sigma(W)$ that we use to regularize each layer of the discriminator is the largest singular value of $W$.
If we naively apply singular value decomposition to compute the $\sigma(W)$ at each round of the algorithm, the algorithm can become computationally heavy. Instead, we can use the power iteration method to estimate $\sigma(W)$~\cite[]{golub2000eigenvalue, yoshida2017spectral}.
With power iteration method, we can estimate the spectral norm with very small additional computational time relative to the full computational cost of the vanilla GANs.
Please see Appendix~\ref{apdsec:algo_snorm} for the detail method and Algorithm ~\ref{algo:snorm} for the summary of the actual spectral normalization algorithm.

\subsection{Gradient analysis of the spectrally normalized weights \label{subsec:grad}}
The gradient\footnote{Indeed, when the spectrum has multiplicities, we would be looking at subgradients here. However, the probability of this happening is zero (almost surely), so we would continue discussions without giving considerations to such events.} of $\bar{W}_{\rm SN}(W)$ with respect to $W_{ij}$ is:
\begin{align}
	\frac{\partial \bar{W}_{\rm SN}(W)}{\partial W_{ij}} = \frac{1}{\sigma(W)}E_{ij} - \frac{1}{\sigma(W)^2} \frac{\partial \sigma(W)}{\partial W_{ij}} W &=\frac{1}{\sigma(W)}E_{ij} - \frac{[\bmu_{1}\bmv_{1}^{\rm T}]_{ij}} {\sigma(W)^2} W \\
    &= \frac{1}{\sigma(W)}\left(E_{ij} - [\bmu_{1}\bmv_{1}^{\rm T}]_{ij} \bar{W}_{\rm SN}\right),
\end{align}
where $E_{ij}$ is the matrix whose $(i,j)$-th entry is $1$ and zero everywhere else, and $\bmu_1$ and $\bmv_1$ are respectively the first left and right singular vectors of $W$.
If $\bmh$ is the hidden layer in the network to be transformed by $\bar{W}_{SN}$, the derivative of the $V(G,D)$ calculated over the mini-batch with respect to $W$ of the discriminator $D$ is given by:
\begin{align}
 \frac{\partial V(G,D)}{\partial W} &= \frac{1}{\sigma(W)} \left(\hat{\E}\left[\bmdelta \bmh^{\rm T}\right] - \left(\hat{\E}\left[\bmdelta^{\rm T} \bar{W}_{\rm SN}\bmh \right] \right) \bmu_1 \bmv_1^{\rm T} \right) \\
 &=\frac{1}{\sigma(W)} \left(\hat{\E}\left[\bmdelta \bmh^{\rm T}\right] - \lambda \bmu_1 \bmv_1^{\rm T} \right) \label{eq:grad_sn}
\end{align}
where $\bmdelta := \left(\partial V(G,D) / \partial \left(\bar{W}_{\rm SN} \bmh\right)\right)^{\rm T}$, $\lambda:=\hat{\E}\left[\bmdelta^{\rm T} \left( \bar{W}_{\rm SN}\bmh \right)\right]$, and $\hat{\E}[\cdot]$ represents empirical expectation over the mini-batch.
$\frac{\partial V}{\partial W}=0$ when $\hat{\E}[\bmdelta \bmh^{\rm T}] = k\bmu_1 \bmv_1^T \label{eq:fixedpoint}$ for some $k\in \bbR$.

We would like to comment on the implication of~\eqref{eq:grad_sn}.
The first term $\hat{\E}\left[\bmdelta \bmh^{\rm T}\right]$ is the same as the derivative of the weights without normalization.
In this light, the second term in the expression can be seen as the regularization term penalizing the first singular components with the \textit{adaptive} regularization coefficient $\lambda$.
$\lambda$ is positive when $\bmdelta$ and $\bar{W}_{\rm SN} \bmh$ are pointing in similar direction, and this prevents the column space of $W$ from concentrating into one particular direction in the course of the training.
In other words, spectral normalization prevents the transformation of each layer from becoming to sensitive in one direction.
We can also use spectral normalization to devise a new parametrization for the model. 
Namely, we can split the layer map into two separate trainable components: spectrally normalized map and the spectral norm constant. As it turns out, this parametrization has its merit on its own and promotes the performance of GANs (See Appendix~\ref{subsec:reparametrization}).
\section{Spectral Normalization vs Other Regularization Techniques}\label{sec:snvsothers}
The weight normalization introduced by~\cite{salimans2016weight}
is a method that normalizes the $\ell_2$ norm of each row vector in the weight matrix. Mathematically, this is equivalent to requiring
the weight by the weight normalization $\bar{W}_{\rm WN}$:
\begin{align}
	\sigma_1(\bar{W}_{\rm WN})^2 + \sigma_2(\bar{W}_{\rm WN})^2 + \cdots + \sigma_T(\bar{W}_{\rm WN})^2 = d_o,~{\rm where}~T=\min(d_i,d_o), \label{eq:svconstraint}
\end{align}
where $\sigma_t(A)$ is a $t$-th singular value of matrix $A$. Therefore, up to a scaler, this is same as the Frobenius normalization, which requires the sum of the squared singular values to be 1.
These normalizations, however, inadvertently impose much stronger constraint on the matrix than intended.
If $\bar{W}_{\rm WN}$ is the weight normalized matrix of dimension $d_i \times d_o$, the norm $\|\bar{W}_{\rm WN}\bmh\|_2$ for a fixed unit vector $\bmh$ is maximized at $\| \bar{W}_{\rm WN} \bmh \|_2 =\sqrt{d_o}$ when $\sigma_1(\bar{W}_{\rm WN})=\sqrt{d_o}$ and $\sigma_t(\bar{W}_{\rm WN}) = 0$ for $t=2,\dots,T$, which means that  $\bar{W}_{\rm WN}$ is of rank one.
Similar thing can be said to the Frobenius normalization (See the appendix for more details).
Using such $\bar{W}_{\rm WN}$ corresponds to using only one feature to discriminate the model probability distribution from the target.
In order to retain as much norm of the input as possible and hence to make the discriminator more sensitive, one would hope to make the norm of $\bar{W}_{\rm WN}\bmh$ large.
For weight normalization, however, this comes at the cost of reducing the rank and hence the number of features to be used for the discriminator.
Thus, there is a conflict of interests between weight normalization and our desire to use as many features as possible to distinguish the generator distribution from the target distribution.
The former interest often reigns over the other in many cases, inadvertently diminishing the number of features to be used by the discriminators.
Consequently, the algorithm would produce a rather arbitrary model distribution that matches the target distribution only at select few features.
Weight clipping~\cite[]{arjovsky2017wasserstein} also suffers from same pitfall.

Our spectral normalization, on the other hand, do not suffer from such a conflict in interest.
Note that the Lipschitz constant of a linear operator is determined only by the maximum singular value.
In other words, the spectral norm is independent of rank.
Thus, unlike the weight normalization, our spectral normalization allows the parameter matrix to use as many features as possible while satisfying local $1$-Lipschitz constraint.
Our spectral normalization leaves more freedom in choosing the number of singular components (features) to feed to the next layer of the discriminator.

\cite{brock2016neural} introduced orthonormal regularization on each weight to stabilize the training of GANs.
In their work, \cite{brock2016neural} augmented the adversarial objective function by adding the following term:
\begin{align}
	\|W^{\rm T} W-I\|_F^2.
\end{align}
While this seems to serve the same purpose as spectral normalization, orthonormal regularization are mathematically quite different from our spectral normalization because the orthonormal regularization destroys the information about the spectrum by setting all the singular values to one. 
On the other hand, spectral normalization only scales the spectrum so that the its maximum will be one. 

\cite{gulrajani2017improved} used Gradient penalty method in combination with WGAN.
In their work, they placed $K$-Lipschitz constant on the discriminator by augmenting the objective function with the regularizer that rewards the function for having local 1-Lipschitz constant (i.e. $\|\nabla_{\hat{\bmx}}f\|_2=1$)  at discrete sets of points of the form $\hat{\bmx} := \epsilon \tilde \bmx + (1- \epsilon) \bmx$ generated by interpolating a sample $\tilde \bmx$ from generative distribution and a sample $\bmx$ from the data distribution.
While this rather straightforward approach does not suffer from the problems we mentioned above regarding the effective dimension of the feature space, the approach has an obvious weakness of being heavily dependent on the support of the current generative distribution.
As a matter of course, the generative distribution and its support gradually changes in the course of the training, and this can destabilize the effect of such regularization.
In fact, we empirically observed that a high learning rate can destabilize the performance of WGAN-GP.
On the contrary, our spectral normalization regularizes the function the operator space, and the effect of the regularization is more stable with respect to the choice of the batch.
Training with our spectral normalization does not easily destabilize with aggressive learning rate.
Moreover, WGAN-GP requires more computational cost than our spectral normalization with single-step power iteration, because the computation of  $\|\nabla_{\hat{\bmx}}f\|_2$ requires one whole round of forward and backward propagation.
In the appendix section, we compare the computational cost of the two methods for the same number of updates.
\section{Experiments}
In order to evaluate the efficacy of our approach and investigate the reason behind its efficacy, we conducted a set of extensive experiments of unsupervised image generation on CIFAR-10~\cite[]{torralba200880} and STL-10~\cite[]{coates2011analysis}, and compared our method against other normalization techniques.
To see how our method fares against large dataset, we also applied our method on ILSVRC2012 dataset (ImageNet)~\cite[]{ILSVRC15} as well.
This section is structured as follows.
First, we will discuss the objective functions we used to train the architecture, and then we will describe the optimization settings we used in the experiments.
We will then explain two performance measures on the images to evaluate the images produced by the trained generators.
Finally, we will summarize our results on CIFAR-10, STL-10, and ImageNet.

As for the architecture of the discriminator and generator, we used convolutional neural networks.
Also, for the evaluation of the spectral norm for the convolutional weight $W \in \bbR^{d_{\rm out}\times d_{\rm in} \times h \times w}$, we treated the operator as a 2-D matrix of dimension $d_{\rm out}\times (d_{\rm in} h w)$\footnote{Note that, since we are conducting the convolution discretely, the spectral norm will depend on the size of the stride and padding. 
However, the answer will only differ by some predefined $K$.}.
We trained the parameters of the generator with batch normalization~\cite[]{ioffe2015batch}.
We refer the readers to Table~\ref{tab:cnn_models} in the appendix section for more details of the architectures.

For all methods other than WGAN-GP, we used the following standard objective function for the adversarial loss:
\begin{align}
	V(G, D) := \E_{x\sim q_{\rm data}(\bmx)} [ \log D(\bmx)] +  \E_{\bmz\sim p(\bmz)} [\log(1-D(G(\bmz)))], \label{eq:advloss}
\end{align}
where $\bmz \in \bbR^{d_z}$ is a latent variable, $p(\bmz)$ is the standard normal distribution $\mathcal{N}(0, I)$, and $G: \bbR^{d_z} \to \bbR^{d_0}$ is a deterministic generator function.
We set $d_z$ to 128 for all of our experiments.
For the updates of $G$, we used the alternate cost proposed by~\cite{goodfellow2014generative}
$-\E_{\bmz\sim p(\bmz)} [\log(D(G(\bmz)))]$
as used in~\cite{goodfellow2014generative} and~\cite{warde2016improving}.
For the updates of $D$, we used the original cost defined in~\eqref{eq:advloss}.
We also tested the performance of the algorithm with the so-called hinge loss, which is given by
\begin{align}
	 V_D(\hat{G}, D) &= \E_{\bmx\sim q_{\rm data}(\bmx)}\left[{\rm min}\left(0, -1+D(\bmx)\right)\right] + \E_{\bmz\sim p(\bmz)} \left[{\rm min}\left(0, -1-D\left(\hat{G}(\bmz)\right)\right)\right]\\
     V_G(G, \hat{D}) &= -\E_{\bmz\sim p(\bmz)}\left[\hat{D}\left(G(\bmz)\right)\right], \label{eq:hinge}
\end{align}
respectively for the discriminator and the generator.
Optimizing these objectives is equivalent to minimizing the so-called reverse KL divergence : ${\rm KL}[p_g||q_{\rm data}]$.
This type of loss has been already proposed and used in~\cite{lim2017geometric, tran2017deep}.
The algorithm based on the hinge loss also showed good performance when evaluated with inception score and FID.
For Wasserstein GANs with gradient penalty (WGAN-GP)~\cite[]{gulrajani2017improved}, we used the following objective function:
$V(G, D) := \E_{\bmx \sim q_{\rm data}} [ D(\bmx)] -  \E_{\bmz\sim p(\bmz)} [D(G(\bmz))] - \lambda \E_{\hat{\bmx} \sim p_{\hat{\bmx}}}[(\|\nabla_{\hat{\bmx}}D(\hat{\bmx})\|_2-1)^2] \label{eq:wgan_gp}$,
where the regularization term is the one we introduced in the appendix section~\ref{subsubsec:wgan-gp}.

For quantitative assessment of generated examples, we used \textit{inception score}~\cite[]{salimans2016improved} and \textit{Fr\'echet inception distance} (FID)~\cite[]{heusel2017gans}.
Please see Appendix~\ref{apdsubsec:performance_measures} for the details of each score.
\subsection{Results on CIFAR10 and STL-10}
In this section, we report the accuracy of the spectral normalization (we use the abbreviation: SN-GAN for the spectrally normalized GANs) during the training, and the dependence of the algorithm's performance on the hyperparmeters of the optimizer.
We also compare the performance quality of the algorithm against those of other regularization/normalization techniques for the discriminator networks, including:
Weight clipping~\cite[]{arjovsky2017wasserstein}, WGAN-GP~\cite[]{gulrajani2017improved}, batch-normalization (BN)~\cite[]{ioffe2015batch}, layer normalization (LN)~\cite[]{ba2016layer}, weight normalization (WN)~\cite[]{salimans2016weight} and orthonormal regularization (\textit{orthonormal})~\cite[]{brock2016neural}.
In order to evaluate the stand-alone efficacy of the gradient penalty, we also applied the gradient penalty term to the standard adversarial loss of GANs~\eqref{eq:advloss}.
We would refer to this method as `GAN-GP'.
For weight clipping, we followed the original work \cite{arjovsky2017wasserstein} and set the clipping constant $c$ at 0.01 for the convolutional weight of each layer.
For gradient penalty, we set $\lambda$ to 10, as suggested in~\cite{gulrajani2017improved}.
For \textit{orthonormal}, we initialized the each weight of $D$ with a randomly selected orthonormal operator and trained GANs with the objective function augmented with the regularization term used in \cite{brock2016neural}.
For all comparative studies throughout, we excluded the multiplier parameter $\bmgamma$ in the weight normalization method, as well as in batch normalization and layer normalization method.
This was done in order to prevent the methods from overtly violating the Lipschitz condition.
When we experimented \textit{with} different multiplier parameter, we were in fact not able to achieve any improvement.

For optimization, we used the Adam optimizer~\cite{kingma2014adam} in all of our experiments.
We tested with 6 settings for (1) $n_{\rm dis}$, the number of updates of the discriminator per one update of the generator and (2) learning rate $\alpha$ and the first and second order momentum parameters ($\beta_1$, $\beta_2$) of Adam.
We list the details of these settings in Table~\ref{tab:settings} in the appendix section.
Out of these 6 settings, A, B, and C are the settings used in previous representative works.
The purpose of the settings D, E, and F is to the evaluate the performance of the algorithms implemented with more aggressive learning rates. For the details of the architectures of convolutional networks deployed in the generator and the discriminator, we refer the readers to Table~\ref{tab:cnn_models} in the appendix section.
The number of updates for GAN generator were 100K for all experiments, unless otherwise noted.

\begin{table}[t]
  \tblcaption{\label{tab:settings}
  Hyper-parameter settings we tested in our experiments. $\dagger$, $\ddagger$ and $\star$ are the hyperparameter settings following~\cite{gulrajani2017improved},~\cite{warde2016improving} and~\cite{radford2015unsupervised}, respectively.}
  \centering
  \small{
  \begin{tabular}{lrrrr}
    \toprule
    \textbf{Setting} & $\alpha$ & $\beta_1$ & $\beta_2$ & $n_{\rm dis}$\\
    \midrule
    A\textsuperscript{$\dagger$}& 0.0001 & 0.5 & 0.9 & 5\\
    B\textsuperscript{$\ddagger$}& 0.0001 & 0.5 & 0.999 & 1\\
    C\textsuperscript{$\star$}& 0.0002 & 0.5 & 0.999 & 1\\
    D & 0.001 & 0.5 & 0.9 & 5\\
    E & 0.001 & 0.5 & 0.999 & 5\\
    F & 0.001 & 0.9 & 0.999 & 5\\
  \bottomrule
  \end{tabular}
  }
\end{table}
Firstly, we inspected the spectral norm of each layer during the training to make sure that our spectral normalization procedure is indeed serving its purpose.
As we can see in the Figure \ref{fig:snorm} in the \ref{apdsubsec:acc_sn}, the spectral norms of these layers floats around $1$--$1.05$ region throughout the training.
Please see Appendix \ref{apdsubsec:acc_sn} for more details.

\begin{figure}[t]
	\centering
    \hspace*{\fill}
    \begin{subfigure}{0.55\textwidth}
		\includegraphics[width=1.0\textwidth]{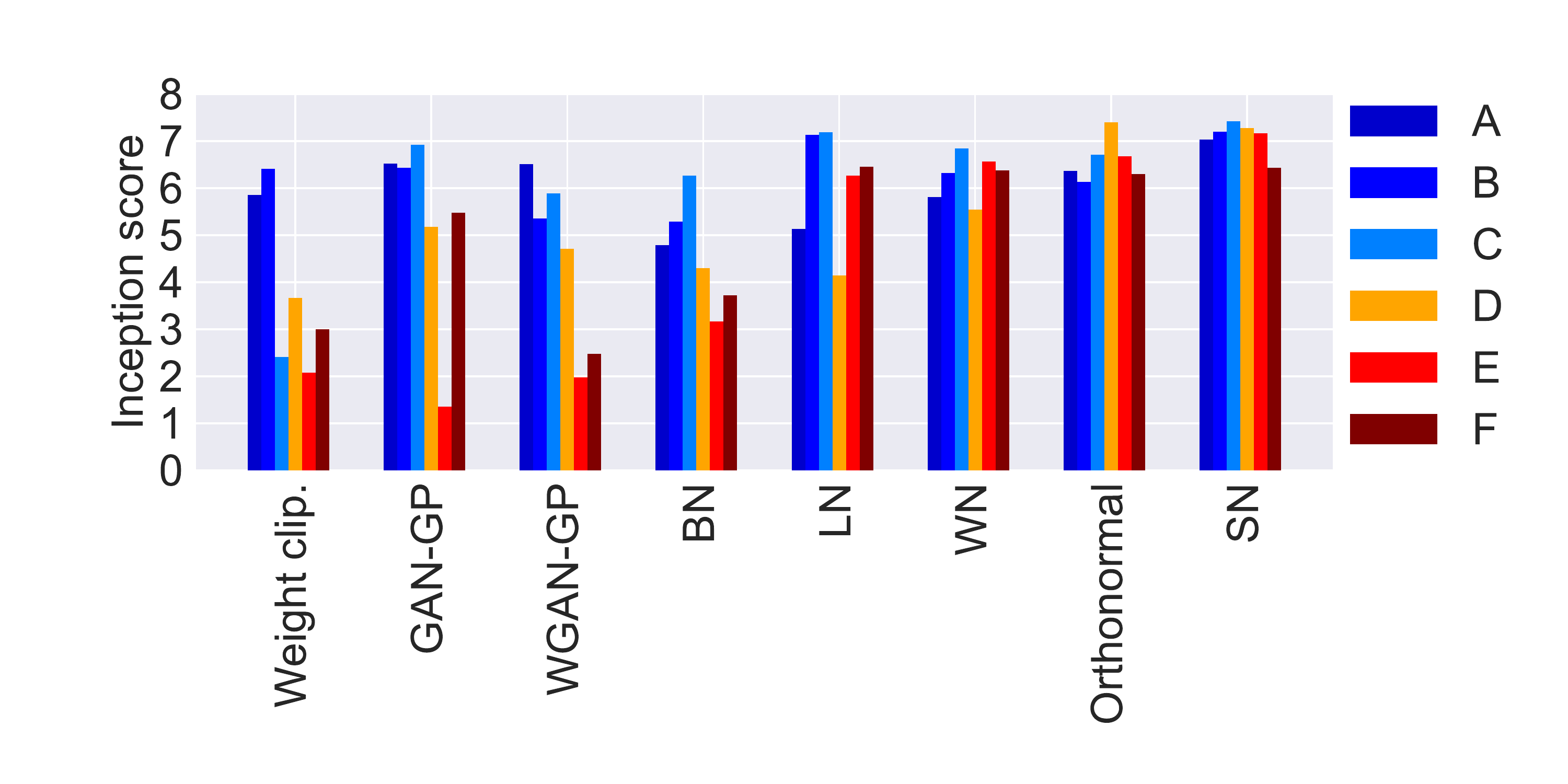}
        \caption{\label{fig:incpscores_cifar10} CIFAR-10}
    \end{subfigure}\hfill%
    \begin{subfigure}{0.4\textwidth}
		\includegraphics[width=1.0\textwidth]
{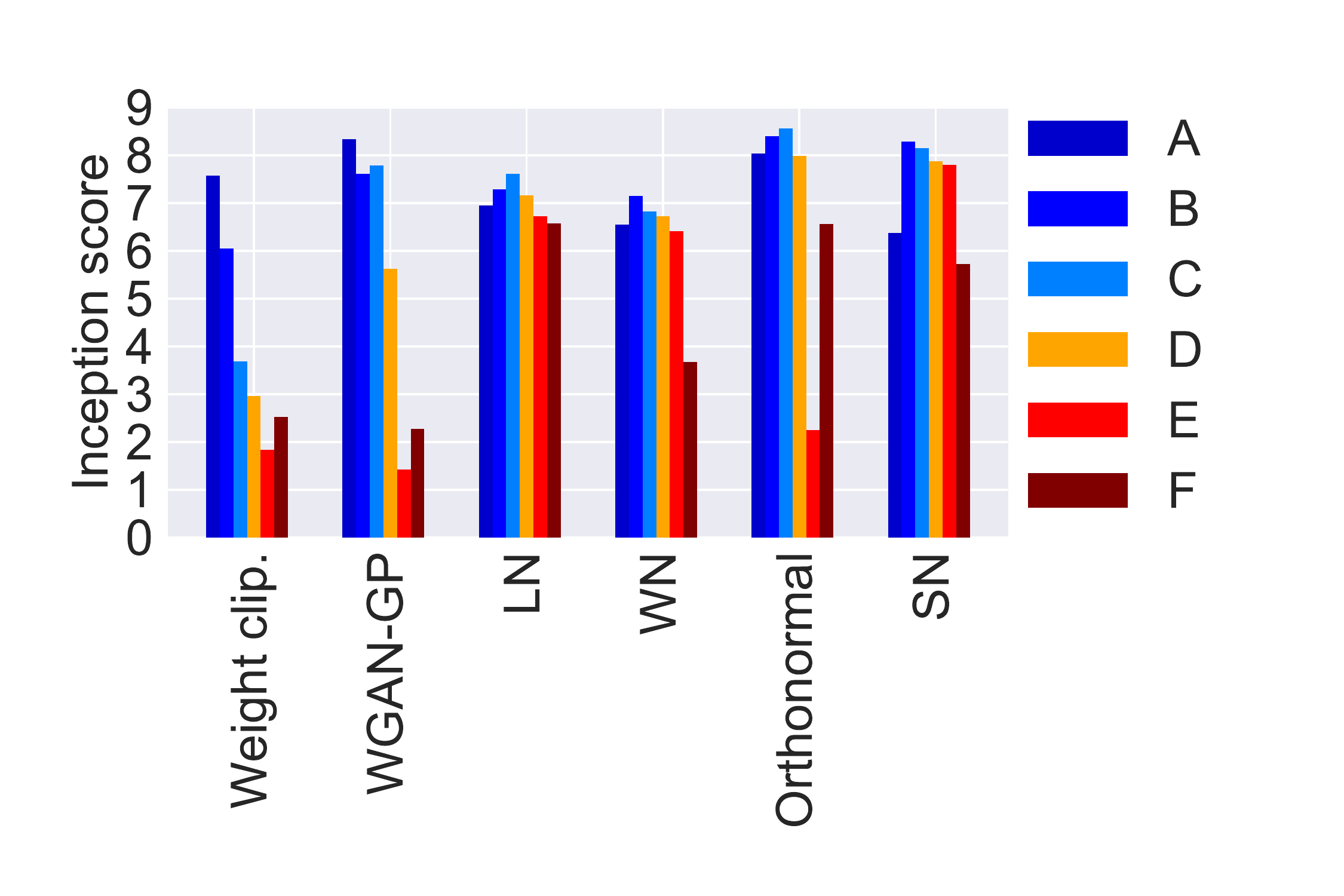}
        \caption{\label{fig:incpscores_stl} STL-10}
    \end{subfigure}
    \hspace*{\fill}%
    \caption{\label{fig:incpscores} Inception scores on CIFAR-10 and STL-10 with different methods and hyperparameters (higher is better). }
\end{figure}

\begin{figure}[t]
	\centering
    \hspace*{\fill}
    \begin{subfigure}{0.55\textwidth}
		\includegraphics[width=1.0\textwidth]{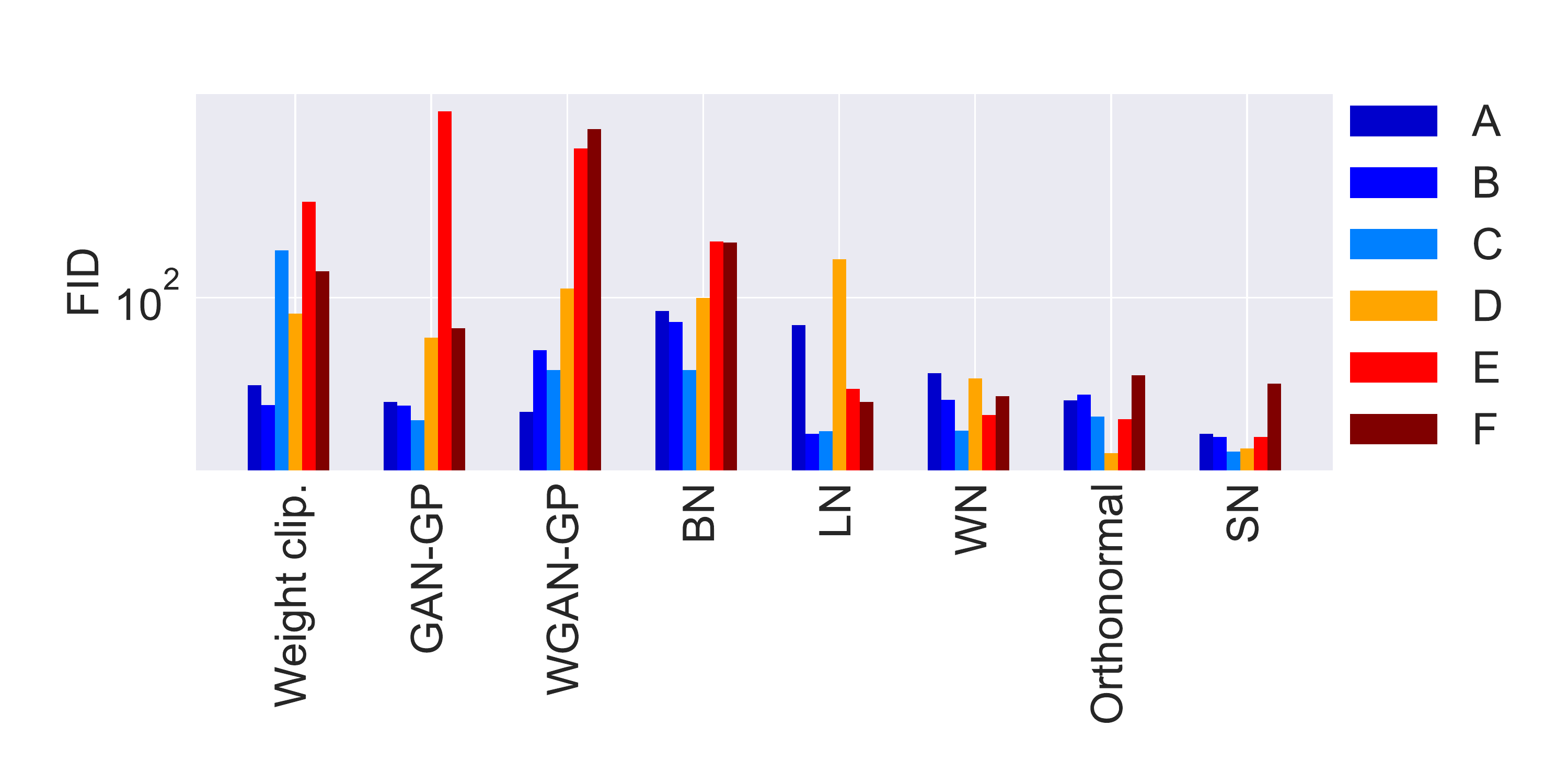}
        \caption{\label{fig:fids_cifar10} CIFAR-10}
    \end{subfigure}\hfill%
    \begin{subfigure}{0.4\textwidth}
		\includegraphics[width=1.0\textwidth]
{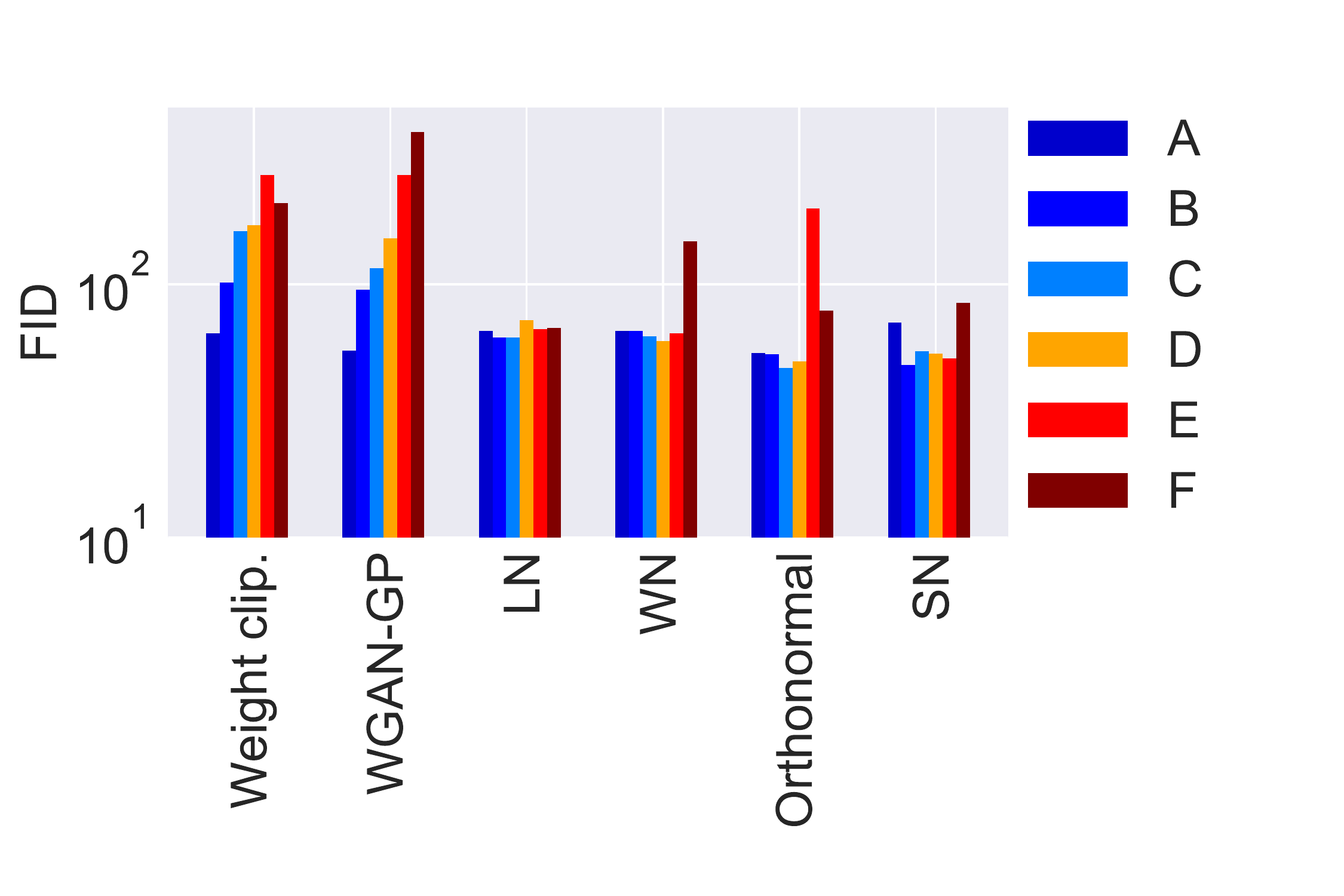}
        \caption{\label{fig:fids_stl} STL-10}
    \end{subfigure}
    \hspace*{\fill}%
    \caption{\label{fig:fids} FIDs on CIFAR-10 and STL-10 with different methods and hyperparameters (lower is better). }
\end{figure}

In Figures~\ref{fig:incpscores} and ~\ref{fig:fids} we show the inception scores of each method with the settings A--F.
We can see that spectral normalization is relatively robust with aggressive learning rates and momentum parameters.
WGAN-GP fails to train good GANs at high learning rates and high momentum parameters on both CIFAR-10 and STL-10. 
Orthonormal regularization performed poorly for the setting E on the STL-10, but performed slightly better than our method with the optimal setting. These results suggests that our method is more robust than other methods with respect to the change in the setting of the training.
Also, the optimal performance of weight normalization was inferior to both WGAN-GP and spectral normalization on STL-10, which consists of more diverse examples than CIFAR-10.
Best scores of spectral normalization are better than almost all other methods on both CIFAR-10 and STL-10.

In Tables~\ref{tab:incepscores}, we show the inception scores of the different methods with optimal settings on CIFAR-10 and STL-10 dataset.
We see that SN-GANs performed better than almost all contemporaries on the optimal settings. SN-GANs performed even better with hinge loss~\eqref{eq:hinge}.\footnote{As for STL-10, we also ran SN-GANs over twice time longer iterations because it did not seem to converge. 
Yet still, this elongated training sequence still completes before WGAN-GP with original iteration size because the optimal setting of SN-GANs (setting B, $n_{dis} =1$) is computationally light.}.
For the training with same number of iterations, SN-GANs fell behind orthonormal regularization for STL-10. For more detailed comparison between orthonormal regularization and spectral normalization, please see section \ref{sec:compare_agstORTH}. 

\begin{table}[t]
\centering
  \caption{\label{tab:incepscores}Inception scores and FIDs with unsupervised image generation on CIFAR-10.
  $\dagger$~\cite[]{radford2015unsupervised} (experimented by ~\cite{yang2017lr}),
  $\ddagger$~\cite[]{yang2017lr}, $*$~\cite[]{warde2016improving},  $\dagger \dagger$~\cite[]{gulrajani2017improved}
  }
  \small{
  \begin{tabular}[t]{lrrrr}
    \toprule
    \multirow{2}{*}{Method} & \multicolumn{2}{c}{\textbf{Inception score}} & \multicolumn{2}{c}{\textbf{FID}} \\
    & CIFAR-10 & STL-10 & CIFAR-10 & STL-10\\
    \midrule
    Real data & 11.24$\pm$.12 & 26.08$\pm$.26 & 7.8 &7.9 \\
    \midrule
    \textbf{-Standard CNN-} \\
    Weight clipping & 6.41$\pm$.11 & 7.57$\pm$.10 & 42.6 & 64.2\\  
    GAN-GP & 6.93$\pm$.08 & & 37.7 & \\
    WGAN-GP & 6.68$\pm$.06 & 8.42$\pm$.13 & 40.2 &  55.1\\
    Batch Norm. & 6.27$\pm$.10 & & 56.3 & \\
    Layer Norm. & 7.19$\pm$.12 & 7.61$\pm$.12  & 33.9 & 75.6\\
    Weight Norm. & 6.84$\pm$.07 &7.16$\pm$.10 & 34.7 &73.4 \\
    Orthonormal & 7.40$\pm$.12& 8.56$\pm$.07 & 29.0 & 46.7 \\
    (ours) SN-GANs & 7.42$\pm$.08 &  8.28$\pm$.09 & 29.3 & 53.1 \\
    Orthonormal (2x updates) & & 8.67$\pm$.08 & & 44.2 \\
    (ours) SN-GANs (2x updates) & & 8.69$\pm$.09 & & 47.5 \\
    \midrule
    (ours) SN-GANs, Eq.\eqref{eq:hinge}  & 7.58$\pm$.12 &  & 25.5 & \\
    (ours) SN-GANs, Eq.\eqref{eq:hinge} (2x updates) && 8.79$\pm$.14 && 43.2 \\
  	\midrule
    \textbf{-ResNet-}\footnotemark \\
    Orthonormal, Eq.\eqref{eq:hinge} & 7.92$\pm$.04  & 8.72$\pm$.06 & 23.8$\pm$.58 &42.4$\pm$.99\\
    (ours) SN-GANs, Eq.\eqref{eq:hinge} &\textbf{8.22}$\pm$.05 & \textbf{9.10}$\pm$.04 & \textbf{21.7}$\pm$.21 & \textbf{40.1}$\pm$.50\\
    \midrule
    DCGAN\textsuperscript{$\dagger$} & 6.64$\pm$.14 & 7.84$\pm$.07\\
    LR-GANs\textsuperscript{$\ddagger$} &  7.17$\pm$.07\\
    Warde-Farley et al.\textsuperscript{$*$} & 7.72$\pm$.13 & 8.51$\pm$.13 \\
    WGAN-GP (ResNet)\textsuperscript{$\dagger \dagger$} & 7.86$\pm$.08 &&& \\
    \bottomrule
  \end{tabular}
  }
\end{table}

In Figure~\ref{fig:images} we show the images produced by the generators trained with WGAN-GP, weight normalization, and spectral normalization.
SN-GANs were consistently better than GANs with weight normalization in terms of the quality of generated images.
To be more precise, as we mentioned in Section~\ref{sec:snvsothers}, the set of images generated by spectral normalization was clearer and more diverse than the images produced by the weight normalization.
We can also see that WGAN-GP failed to train good GANs with high learning rates and high momentums (D,E and F).
The generated images with GAN-GP, batch normalization, and layer normalization is shown in Figure~\ref{fig:cifar10images_others} in the appendix section.

We also compared our algorithm against multiple benchmark methods ans summarized the results on the bottom half of the Table~\ref{tab:incepscores}.
We also tested the performance of our method on ResNet based GANs used in \cite{gulrajani2017improved}.
Please note that all methods listed thereof are all different in both optimization methods and the architecture of the model. 
Please see Table~\ref{tab:resnets_cifar10} and ~\ref{tab:resnets_stl} in the appendix section for the detail network architectures.
Our implementation of our algorithm was able to perform better than almost all the predecessors in the performance.  
\footnotetext{For our ResNet experiments, we trained the same architecture with multiple random seeds for weight initialization and produced models with different parameters.
  We then generated 5000 images 10 times and computed the average inception score for each model. 
  The values for ResNet on the table are the mean and standard deviation of the score computed over the set of models trained with different seeds.}
  
\subsubsection{Analysis of SN-GANs}
\begin{figure}[t]
	\centering
    \begin{subfigure}{1.0\textwidth}
		\includegraphics[width=1.0\textwidth]{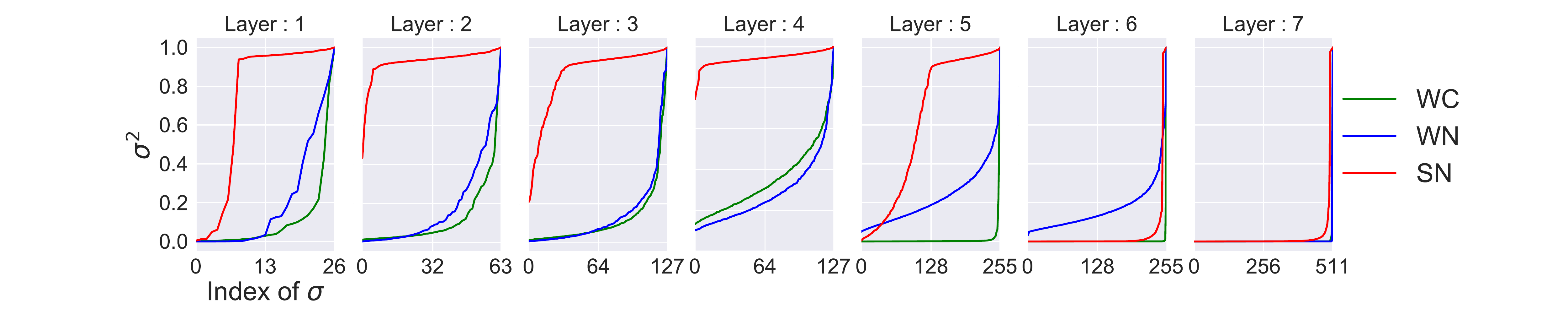}
        \caption{\label{fig:svhist_cifar10} CIFAR-10}
    \end{subfigure}
    \begin{subfigure}{1.0\textwidth}
		\includegraphics[width=1.0\textwidth]{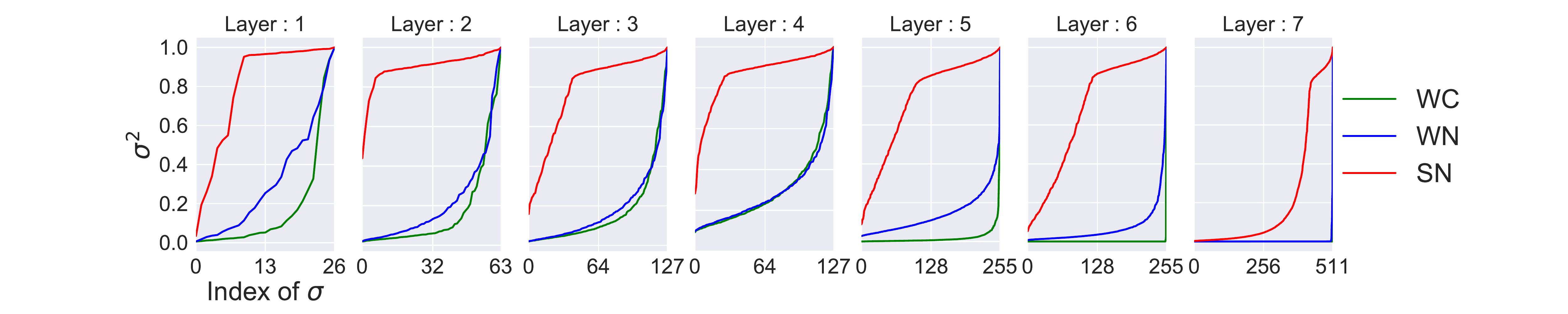}
        \caption{\label{fig:svhist_stl} STL-10}
    \end{subfigure}
    \caption{Squared singular values of weight matrices trained with different methods: Weight clipping~(WC), Weight Normalization~(WN) and Spectral Normalization~(SN).
      We scaled the singular values so that the largest singular values is equal to 1.
      For WN and SN, we calculated singular values of the normalized weight matrices. }\label{fig:svhist}
\end{figure}
\begin{figure}[H]
    \begin{subfigure}{1.0\textwidth}
        \centering
        \includegraphics[width=1.0\textwidth]{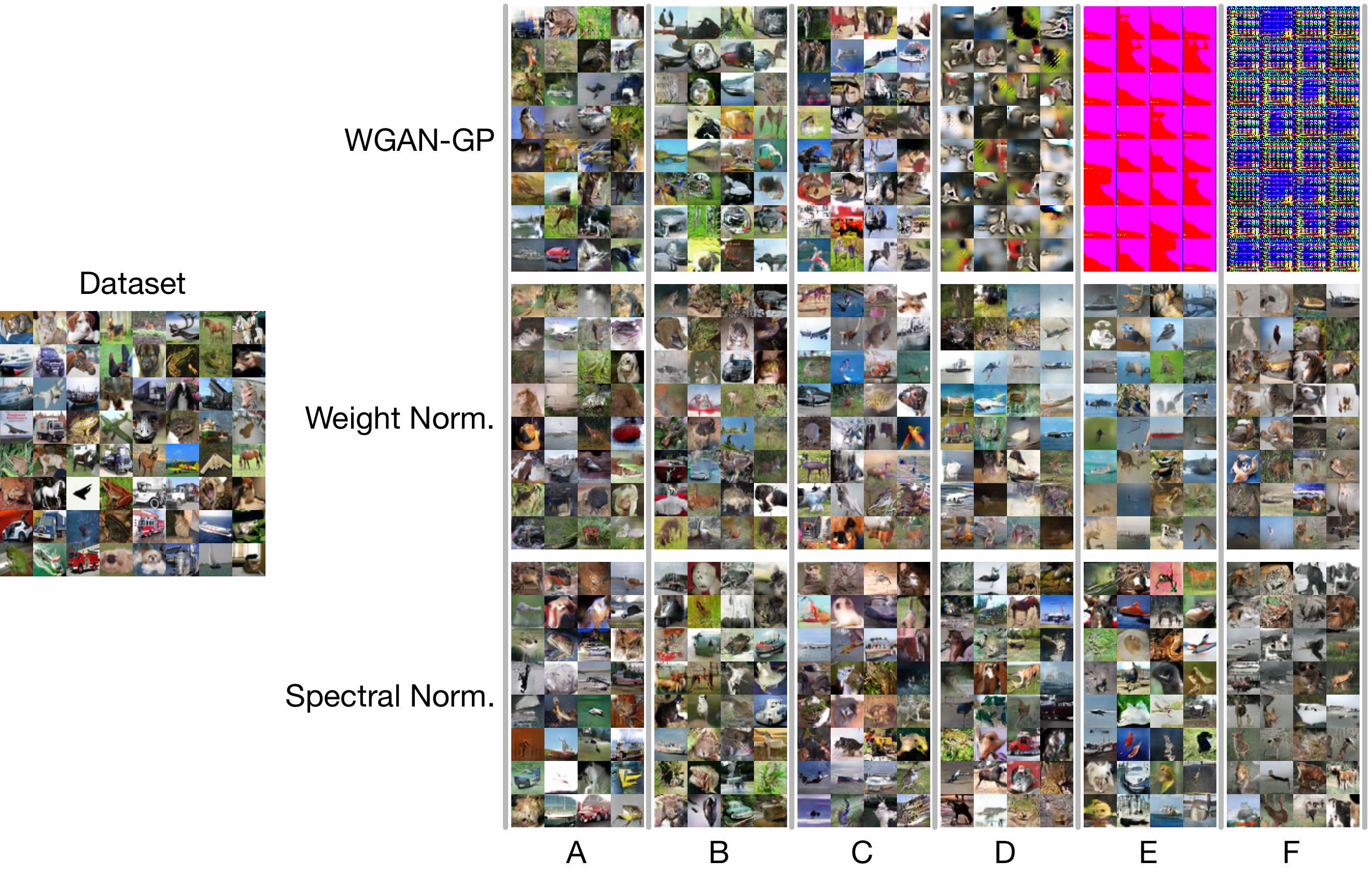}
        \caption{\label{fig:cifar10images} CIFAR-10}
    \end{subfigure}
	\begin{subfigure}{1.0\textwidth}
    	\centering
 		\includegraphics[width=1.0\textwidth]{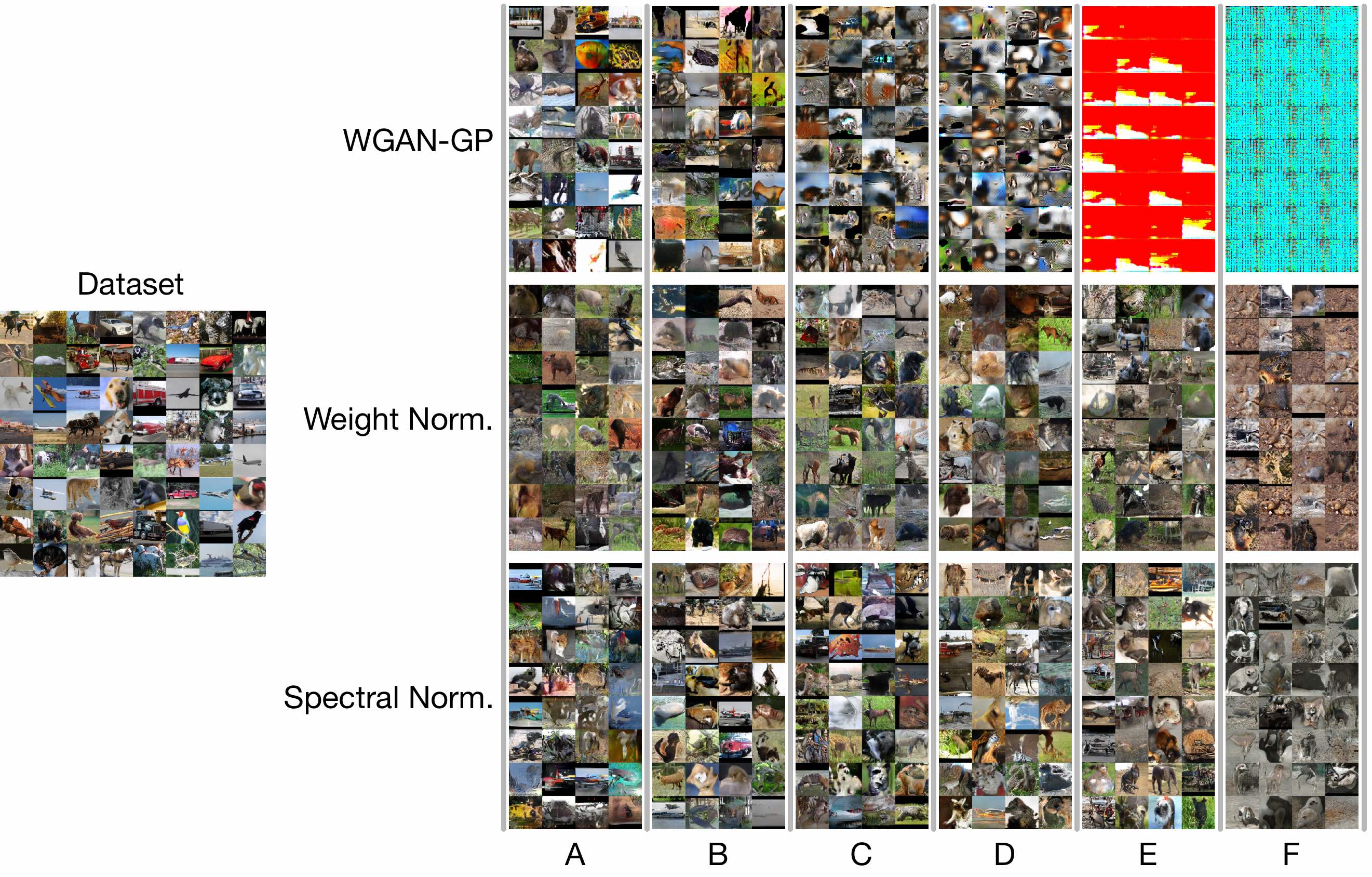}
         \caption{\label{fig:stlimages} STL-10}
    \end{subfigure}
    \caption{\label{fig:images} Generated images on different methods: WGAN-GP, weight normalization, and spectral normalization on CIFAR-10 and STL-10.}
\end{figure}

\paragraph{Singular values analysis on the weights of the discriminator $D$}
In Figure~\ref{fig:svhist}, we show the squared singular values of the weight matrices in the final discriminator $D$ produced by each method using the parameter that yielded the best inception score.
As we predicted in Section~\ref{sec:snvsothers}, the singular values of the first to fifth layers trained with weight clipping and weight normalization concentrate on a few components.
That is, the weight matrices of these layers tend to be rank deficit.
On the other hand, the singular values of the weight matrices in those layers trained with spectral normalization is more broadly distributed.
When the goal is to distinguish a pair of probability distributions on the low-dimensional nonlinear data manifold embedded in a high dimensional space, rank deficiencies in lower layers can be especially fatal.
Outputs of lower layers have gone through only a few sets of rectified linear transformations, which means that they tend to lie on the space that is linear in most parts.
Marginalizing out many features of the input distribution in such space can result in oversimplified discriminator.
We can actually confirm the effect of this phenomenon on the generated images especially in Figure~\ref{fig:stlimages}.
The images generated with spectral normalization is more diverse and complex than those generated with weight normalization.
\paragraph{Training time}
On CIFAR-10, SN-GANs is slightly slower than weight normalization (about 110 $\sim$ 120\% computational time), but significantly faster than WGAN-GP. 
As we mentioned in Section~\ref{sec:snvsothers}, WGAN-GP is slower than other methods because WGAN-GP needs to calculate the gradient of gradient norm $\|\nabla_{\bmx} D\|_2$.
For STL-10, the computational time of SN-GANs is almost the same as vanilla GANs, because the relative computational cost of the power iteration~\eqref{eq:poweriter} is negligible when compared to the cost of forward and backward propagation on CIFAR-10 (images size of STL-10 is larger ($48 \times 48$)).
Please see Figure~\ref{fig:cptime} in the appendix section for the actual computational time.

\subsubsection{Comparison between SN-GANs and orthonormal regularization} \label{sec:compare_agstORTH} 
\begin{figure}[t]
	\centering
    \includegraphics[width=0.4\textwidth]{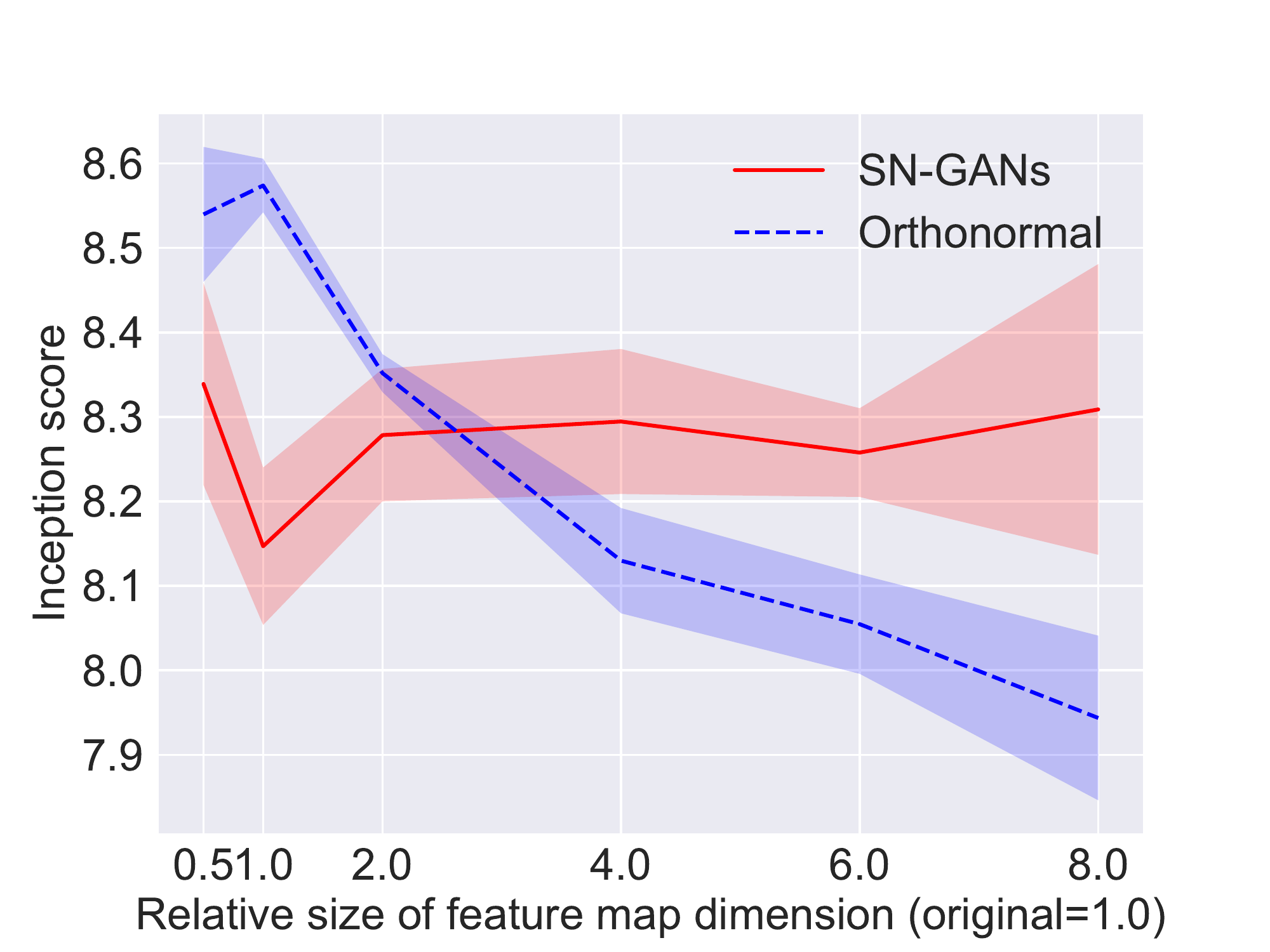}
	\caption{\label{fig:varying_fmaps} The effect on the performance on STL-10 induced by the change of the feature map dimension of the final layer.
    The width of the highlighted region represents standard deviation of the results over multiple seeds of weight initialization.
    The \textit{orthonormal regularization} does not perform well with large feature map dimension, possibly because of its design that forces the discriminator to use all dimensions including the ones that are unnecessary.
For the setting of the optimizers' hyper-parameters, We used the setting C, which was optimal for ``\textit{orthonormal regularization}" }
\end{figure}

In order to highlight the difference between our spectral normalization and orthonormal regularization, we conducted an additional set of experiments.
As we explained in Section~\ref{sec:snvsothers}, orthonormal regularization is different from our method in that it destroys the spectral information and puts equal emphasis on all feature dimensions, including the ones that 'shall' be weeded out in the training process. 
To see the extent of its possibly detrimental effect, we experimented by increasing the dimension of the feature space \footnote{More precisely, we simply increased the input dimension and the output dimension by the same factor.
In Figure ~\ref{fig:varying_fmaps}, `relative size' = 1.0 implies that the layer structure is the same as the original.}, especially at the final layer (7th conv) for which the training with our spectral normalization prefers relatively small feature space (dimension $<100$; see Figure~\ref{fig:svhist_stl}). 
As for the setting of the training, we selected the parameters for which the orthonormal regularization performed optimally.  
The figure~\ref{fig:varying_fmaps} shows the result of our experiments.
As we predicted, the performance of the orthonormal regularization deteriorates as we increase the dimension of the feature maps at the final layer.
Our SN-GANs, on the other hand, does not falter with this modification of the architecture.
Thus, at least in this perspective, we may such that our method is more robust with respect to the change of the network architecture.

\subsection{Image Generation on ImageNet}
% \begin{figure}[t]
%      \centering
%      \includegraphics[width=0.4\textwidth]{figures/imagenet_incp_snall_vs_orthall}
%      \figcaption{\label{fig:imagenet_incp} Learning curves of Inception score with SN-GANs and orthonormal regularization for the conditional image generation task on ImageNet.\\
%      }
% \end{figure}

\begin{figure}[t]
     	\centering
     	\includegraphics[width=0.4\textwidth]{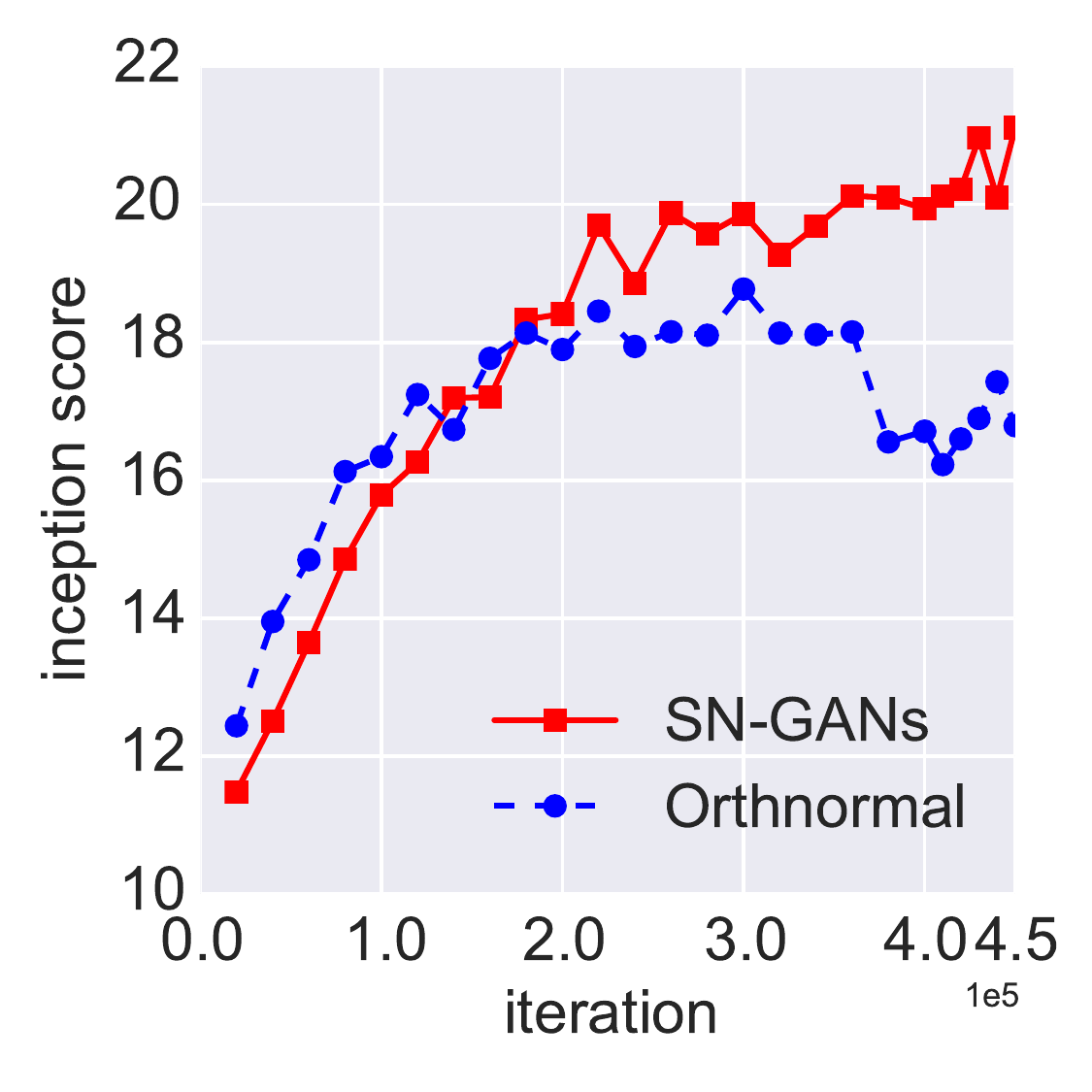}
     \figcaption{\label{fig:imagenet_incp} Learning curves for conditional image generation in terms of Inception score for SN-GANs and GANs with \textit{orthonormal regularization} on ImageNet.}
\end{figure}

To show that our method remains effective on a large high dimensional dataset, we also applied our method to the training of conditional GANs on ILRSVRC2012 dataset with 1000 classes, each consisting of approximately 1300 images, which we compressed to $128 \times 128$ pixels.
Regarding the adversarial loss for conditional GANs, we used practically the same formulation used in \cite{mirza2014conditional}, except that we replaced the standard GANs loss with hinge loss \eqref{eq:hinge}.
Please see Appendix~\ref{apdsubsec:exp_imagenet} for the details of experimental settings.

\begin{figure}[htp]
	\centering
 	\includegraphics[width=1.0\textwidth]{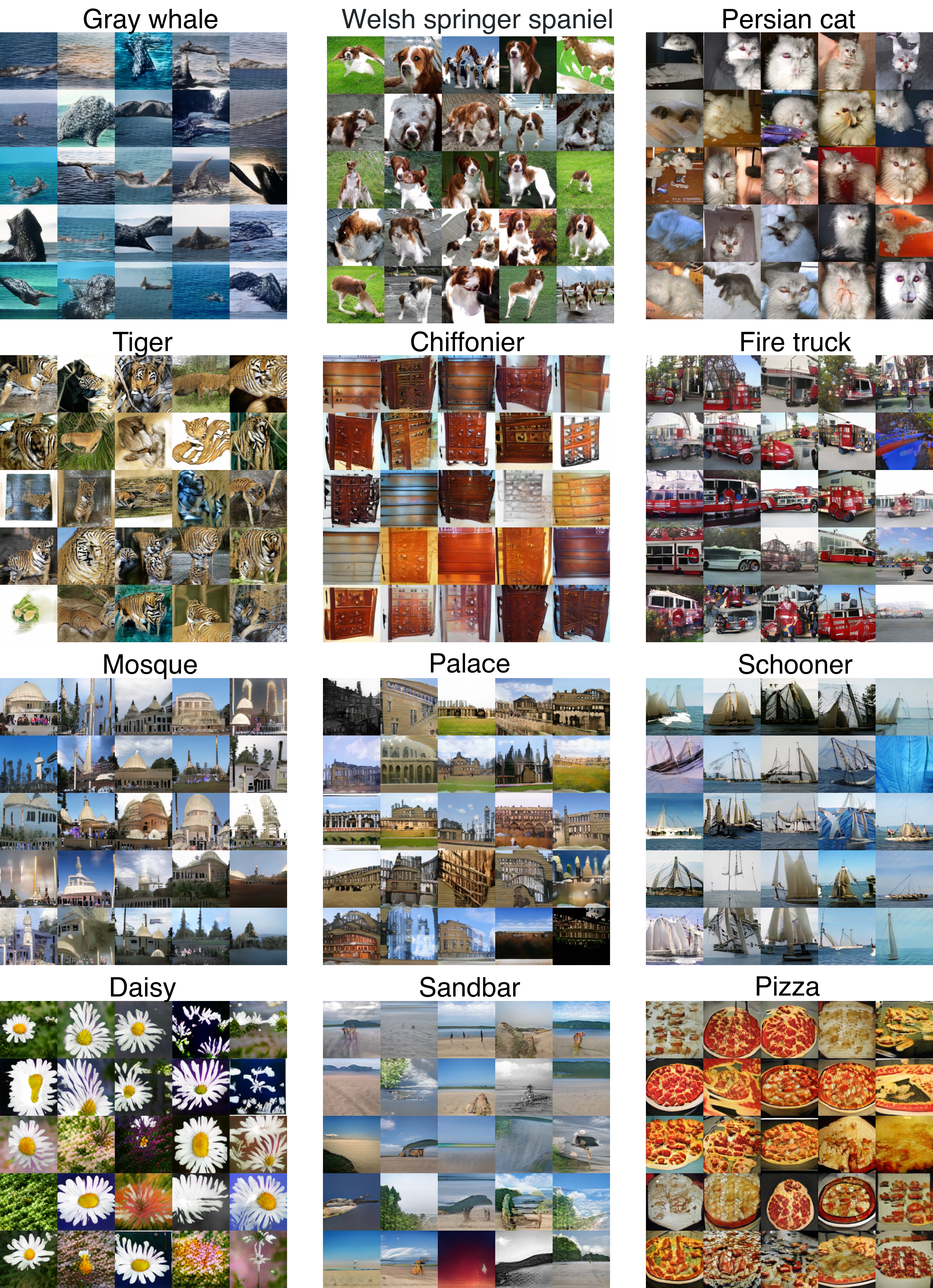}
    \caption{\label{fig:generated_images_imagenet}
    128x128 pixel images generated by SN-GANs trained on ILSVRC2012 dataset.
    The inception score is 21.1$\pm$.35.}
\end{figure}

GANs without normalization and GANs with layer normalization collapsed in the beginning of training and failed to produce any meaningful images. 
GANs with orthonormal normalization \cite{brock2016neural} and our spectral normalization, on the other hand, was able to produce images. 
The inception score of the orthonormal normalization however plateaued around 20Kth iterations, while SN kept improving even afterward (Figure~\ref{fig:imagenet_incp}.) 
To our knowledge, our research is the first of its kind in succeeding to produce decent images from ImageNet dataset with a \textit{single} pair of a discriminator and a generator (Figure~\ref{fig:generated_images_imagenet}).
To measure the degree of mode-collapse, we followed the footstep of~\cite{odena2016conditional} and computed the intra MS-SSIM~\cite{odena2016conditional} for pairs of independently generated GANs images of each class.
We see that our SN-GANs ((intra MS-SSIM)=0.101) is suffering less from the mode-collapse than AC-GANs ((intra MS-SSIM)$\sim$0.25).

To ensure that the superiority of our method is not limited within our specific setting, we also compared the performance of SN-GANs against \textit{orthonormal regularization} on conditional GANs with \textit{projection discriminator}~\cite[]{miyato2018cgans} as well as the standard (unconditional) GANs. 
In our experiments, SN-GANs achieved better performance than \textit{orthonormal regularization} for the both settings (See Figure~\ref{fig:imagenet_incp_others} in the appendix section).

\section{Conclusion}
This paper proposes spectral normalization as a stabilizer of training of GANs.
When we apply spectral normalization to the GANs on image generation tasks, the generated examples are more diverse than the conventional weight normalization and achieve better or comparative inception scores relative to previous studies.
The method imposes global regularization on the discriminator as opposed to local regularization introduced by WGAN-GP, and can possibly used in combinations.
In the future work, we would like to further investigate where our methods stand amongst other methods on more theoretical basis, and experiment our algorithm on larger and more complex datasets.

\subsubsection*{Acknowledgments}
 We would like to thank the members of Preferred Networks, Inc., particularly Shin-ichi Maeda, Eiichi Matsumoto, Masaki Watanabe and Keisuke Yahata for insightful comments and discussions.
We also would like to thank anonymous reviewers and commenters on the OpenReview forum for insightful discussions.

\bibliographystyle{iclr2018_conference}
\bibliography{main}

\begin{thebibliography}{44}
\providecommand{\natexlab}[1]{#1}
\providecommand{\url}[1]{\texttt{#1}}
\expandafter\ifx\csname urlstyle\endcsname\relax
  \providecommand{\doi}[1]{doi: #1}\else
  \providecommand{\doi}{doi: \begingroup \urlstyle{rm}\Url}\fi

\bibitem[Arjovsky \& Bottou(2017)Arjovsky and Bottou]{arjovsky2017towards}
Martin Arjovsky and L{\'e}on Bottou.
\newblock Towards principled methods for training generative adversarial
  networks.
\newblock In \emph{ICLR}, 2017.

\bibitem[Arjovsky et~al.(2017)Arjovsky, Chintala, and
  Bottou]{arjovsky2017wasserstein}
Martin Arjovsky, Soumith Chintala, and L{\'e}on Bottou.
\newblock Wasserstein generative adversarial networks.
\newblock In \emph{ICML}, pp.\  214--223, 2017.

\bibitem[Arpit et~al.(2016)Arpit, Zhou, Kota, and
  Govindaraju]{arpit2016normalization}
Devansh Arpit, Yingbo Zhou, Bhargava~U Kota, and Venu Govindaraju.
\newblock Normalization propagation: A parametric technique for removing
  internal covariate shift in deep networks.
\newblock In \emph{ICML}, pp.\  1168--1176, 2016.

\bibitem[Ba et~al.(2016)Ba, Kiros, and Hinton]{ba2016layer}
Jimmy~Lei Ba, Jamie~Ryan Kiros, and Geoffrey~E Hinton.
\newblock Layer normalization.
\newblock \emph{arXiv preprint arXiv:1607.06450}, 2016.

\bibitem[Brock et~al.(2016)Brock, Lim, Ritchie, and Weston]{brock2016neural}
Andrew Brock, Theodore Lim, James~M Ritchie, and Nick Weston.
\newblock Neural photo editing with introspective adversarial networks.
\newblock \emph{arXiv preprint arXiv:1609.07093}, 2016.

\bibitem[Coates et~al.(2011)Coates, Ng, and Lee]{coates2011analysis}
Adam Coates, Andrew Ng, and Honglak Lee.
\newblock An analysis of single-layer networks in unsupervised feature
  learning.
\newblock In \emph{AISTATS}, pp.\  215--223, 2011.

\bibitem[de~Vries et~al.(2017)de~Vries, Strub, Mary, Larochelle, Pietquin, and
  Courville]{de2017modulating}
Harm de~Vries, Florian Strub, J{\'e}r{\'e}mie Mary, Hugo Larochelle, Olivier
  Pietquin, and Aaron~C Courville.
\newblock Modulating early visual processing by language.
\newblock In \emph{NIPS}, pp.\  6576--6586, 2017.

\bibitem[Dowson \& Landau(1982)Dowson and Landau]{dowson1982frechet}
DC~Dowson and BV~Landau.
\newblock The fr{\'e}chet distance between multivariate normal distributions.
\newblock \emph{Journal of Multivariate Analysis}, 12\penalty0 (3):\penalty0
  450--455, 1982.

\bibitem[Dumoulin et~al.(2017)Dumoulin, Shlens, and
  Kudlur]{dumoulin2016learned}
Vincent Dumoulin, Jonathon Shlens, and Manjunath Kudlur.
\newblock A learned representation for artistic style.
\newblock In \emph{ICLR}, 2017.

\bibitem[Glorot et~al.(2011)Glorot, Bordes, and Bengio]{glorot2011deep}
Xavier Glorot, Antoine Bordes, and Yoshua Bengio.
\newblock Deep sparse rectifier neural networks.
\newblock In \emph{AISTATS}, pp.\  315--323, 2011.

\bibitem[Golub \& Van~der Vorst(2000)Golub and Van~der
  Vorst]{golub2000eigenvalue}
Gene~H Golub and Henk~A Van~der Vorst.
\newblock Eigenvalue computation in the 20th century.
\newblock \emph{Journal of Computational and Applied Mathematics}, 123\penalty0
  (1):\penalty0 35--65, 2000.

\bibitem[Goodfellow et~al.(2014)Goodfellow, Pouget-Abadie, Mirza, Xu,
  Warde-Farley, Ozair, Courville, and Bengio]{goodfellow2014generative}
Ian Goodfellow, Jean Pouget-Abadie, Mehdi Mirza, Bing Xu, David Warde-Farley,
  Sherjil Ozair, Aaron Courville, and Yoshua Bengio.
\newblock Generative adversarial nets.
\newblock In \emph{NIPS}, pp.\  2672--2680, 2014.

\bibitem[Gulrajani et~al.(2017)Gulrajani, Ahmed, Arjovsky, Dumoulin, and
  Courville]{gulrajani2017improved}
Ishaan Gulrajani, Faruk Ahmed, Martin Arjovsky, Vincent Dumoulin, and Aaron
  Courville.
\newblock Improved training of wasserstein \rm{GANs}.
\newblock \emph{arXiv preprint arXiv:1704.00028}, 2017.

\bibitem[He et~al.(2016)He, Zhang, Ren, and Sun]{he2016deep}
Kaiming He, Xiangyu Zhang, Shaoqing Ren, and Jian Sun.
\newblock Deep residual learning for image recognition.
\newblock In \emph{CVPR}, pp.\  770--778, 2016.

\bibitem[Heusel et~al.(2017)Heusel, Ramsauer, Unterthiner, Nessler, Klambauer,
  and Hochreiter]{heusel2017gans}
Martin Heusel, Hubert Ramsauer, Thomas Unterthiner, Bernhard Nessler,
  G{\"u}nter Klambauer, and Sepp Hochreiter.
\newblock \rm{GANs} trained by a two time-scale update rule converge to a nash
  equilibrium.
\newblock \emph{arXiv preprint arXiv:1706.08500}, 2017.

\bibitem[Ho \& Ermon(2016)Ho and Ermon]{ho2016generative}
Jonathan Ho and Stefano Ermon.
\newblock Generative adversarial imitation learning.
\newblock In \emph{NIPS}, pp.\  4565--4573, 2016.

\bibitem[Ioffe \& Szegedy(2015)Ioffe and Szegedy]{ioffe2015batch}
Sergey Ioffe and Christian Szegedy.
\newblock Batch normalization: Accelerating deep network training by reducing
  internal covariate shift.
\newblock In \emph{ICML}, pp.\  448--456, 2015.

\bibitem[Jarrett et~al.(2009)Jarrett, Kavukcuoglu, Ranzato, and
  LeCun]{jarrett2009best}
Kevin Jarrett, Koray Kavukcuoglu, Marc'Aurelio Ranzato, and Yann LeCun.
\newblock What is the best multi-stage architecture for object recognition?
\newblock In \emph{ICCV}, pp.\  2146--2153, 2009.

\bibitem[Jia et~al.(2017)Jia, Tao, Gao, and Xu]{jia2016improving}
Kui Jia, Dacheng Tao, Shenghua Gao, and Xiangmin Xu.
\newblock Improving training of deep neural networks via singular value
  bounding.
\newblock In \emph{CVPR}, 2017.

\bibitem[Kingma \& Ba(2015)Kingma and Ba]{kingma2014adam}
Diederik Kingma and Jimmy Ba.
\newblock Adam: A method for stochastic optimization.
\newblock In \emph{ICLR}, 2015.

\bibitem[Li et~al.(2017)Li, Monroe, Shi, Ritter, and
  Jurafsky]{li2017adversarial}
Jiwei Li, Will Monroe, Tianlin Shi, Alan Ritter, and Dan Jurafsky.
\newblock Adversarial learning for neural dialogue generation.
\newblock In \emph{EMNLP}, pp.\  2147--2159, 2017.

\bibitem[Lim \& Ye(2017)Lim and Ye]{lim2017geometric}
Jae~Hyun Lim and Jong~Chul Ye.
\newblock Geometric \rm{GAN}.
\newblock \emph{arXiv preprint arXiv:1705.02894}, 2017.

\bibitem[Maas et~al.(2013)Maas, Hannun, and Ng]{maas2013rectifier}
Andrew~L Maas, Awni~Y Hannun, and Andrew~Y Ng.
\newblock Rectifier nonlinearities improve neural network acoustic models.
\newblock In \emph{ICML Workshop on Deep Learning for Audio, Speech and
  Language Processing}, 2013.

\bibitem[Mirza \& Osindero(2014)Mirza and Osindero]{mirza2014conditional}
Mehdi Mirza and Simon Osindero.
\newblock Conditional generative adversarial nets.
\newblock \emph{arXiv preprint arXiv:1411.1784}, 2014.

\bibitem[Miyato \& Koyama(2018)Miyato and Koyama]{miyato2018cgans}
Takeru Miyato and Masanori Koyama.
\newblock {cGANs} with projection discriminator.
\newblock In \emph{ICLR}, 2018.

\bibitem[Mohamed \& Lakshminarayanan(2017)Mohamed and
  Lakshminarayanan]{mohamed2016learning}
Shakir Mohamed and Balaji Lakshminarayanan.
\newblock Learning in implicit generative models.
\newblock \emph{NIPS Workshop on Adversarial Training}, 2017.

\bibitem[Nair \& Hinton(2010)Nair and Hinton]{nair2010rectified}
Vinod Nair and Geoffrey~E Hinton.
\newblock Rectified linear units improve restricted boltzmann machines.
\newblock In \emph{ICML}, pp.\  807--814, 2010.

\bibitem[Nowozin et~al.(2016)Nowozin, Cseke, and Tomioka]{nowozin2016f}
Sebastian Nowozin, Botond Cseke, and Ryota Tomioka.
\newblock f-\rm{GAN}: Training generative neural samplers using variational
  divergence minimization.
\newblock In \emph{NIPS}, pp.\  271--279, 2016.

\bibitem[Odena et~al.(2017)Odena, Olah, and Shlens]{odena2016conditional}
Augustus Odena, Christopher Olah, and Jonathon Shlens.
\newblock Conditional image synthesis with auxiliary classifier \rm{GANs}.
\newblock In \emph{ICML}, pp.\  2642--2651, 2017.

\bibitem[Qi(2017)]{qi2017loss}
Guo-Jun Qi.
\newblock Loss-sensitive generative adversarial networks on lipschitz
  densities.
\newblock \emph{arXiv preprint arXiv:1701.06264}, 2017.

\bibitem[Radford et~al.(2016)Radford, Metz, and
  Chintala]{radford2015unsupervised}
Alec Radford, Luke Metz, and Soumith Chintala.
\newblock Unsupervised representation learning with deep convolutional
  generative adversarial networks.
\newblock In \emph{ICLR}, 2016.

\bibitem[Russakovsky et~al.(2015)Russakovsky, Deng, Su, Krause, Satheesh, Ma,
  Huang, Karpathy, Khosla, Bernstein, Berg, and Fei-Fei]{ILSVRC15}
Olga Russakovsky, Jia Deng, Hao Su, Jonathan Krause, Sanjeev Satheesh, Sean Ma,
  Zhiheng Huang, Andrej Karpathy, Aditya Khosla, Michael Bernstein,
  Alexander~C. Berg, and Li~Fei-Fei.
\newblock Imagenet large scale visual recognition challenge.
\newblock \emph{International Journal of Computer Vision}, 115\penalty0
  (3):\penalty0 211--252, 2015.

\bibitem[Saito et~al.(2017)Saito, Matsumoto, and Saito]{saito2016temporal}
Masaki Saito, Eiichi Matsumoto, and Shunta Saito.
\newblock Temporal generative adversarial nets with singular value clipping.
\newblock In \emph{ICCV}, 2017.

\bibitem[Salimans \& Kingma(2016)Salimans and Kingma]{salimans2016weight}
Tim Salimans and Diederik~P Kingma.
\newblock Weight normalization: A simple reparameterization to accelerate
  training of deep neural networks.
\newblock In \emph{NIPS}, pp.\  901--909, 2016.

\bibitem[Salimans et~al.(2016)Salimans, Goodfellow, Zaremba, Cheung, Radford,
  and Chen]{salimans2016improved}
Tim Salimans, Ian Goodfellow, Wojciech Zaremba, Vicki Cheung, Alec Radford, and
  Xi~Chen.
\newblock Improved techniques for training \rm{GANs}.
\newblock In \emph{NIPS}, pp.\  2226--2234, 2016.

\bibitem[Szegedy et~al.(2015)Szegedy, Liu, Jia, Sermanet, Reed, Anguelov,
  Erhan, Vanhoucke, and Rabinovich]{szegedy2015going}
Christian Szegedy, Wei Liu, Yangqing Jia, Pierre Sermanet, Scott Reed, Dragomir
  Anguelov, Dumitru Erhan, Vincent Vanhoucke, and Andrew Rabinovich.
\newblock Going deeper with convolutions.
\newblock In \emph{CVPR}, pp.\  1--9, 2015.

\bibitem[Tokui et~al.(2015)Tokui, Oono, Hido, and Clayton]{tokui2015chainer}
Seiya Tokui, Kenta Oono, Shohei Hido, and Justin Clayton.
\newblock Chainer: a next-generation open source framework for deep learning.
\newblock In \emph{Proceedings of workshop on machine learning systems
  (LearningSys) in the twenty-ninth annual conference on neural information
  processing systems (NIPS)}, 2015.

\bibitem[Torralba et~al.(2008)Torralba, Fergus, and Freeman]{torralba200880}
Antonio Torralba, Rob Fergus, and William~T Freeman.
\newblock 80 million tiny images: A large data set for nonparametric object and
  scene recognition.
\newblock \emph{IEEE Transactions on Pattern Analysis and Machine
  Intelligence}, 30\penalty0 (11):\penalty0 1958--1970, 2008.

\bibitem[Tran et~al.(2017)Tran, Ranganath, and Blei]{tran2017deep}
Dustin Tran, Rajesh Ranganath, and David~M Blei.
\newblock Deep and hierarchical implicit models.
\newblock \emph{arXiv preprint arXiv:1702.08896}, 2017.

\bibitem[Uehara et~al.(2016)Uehara, Sato, Suzuki, Nakayama, and
  Matsuo]{uehara2016generative}
Masatoshi Uehara, Issei Sato, Masahiro Suzuki, Kotaro Nakayama, and Yutaka
  Matsuo.
\newblock Generative adversarial nets from a density ratio estimation
  perspective.
\newblock \emph{NIPS Workshop on Adversarial Training}, 2016.

\bibitem[Warde-Farley \& Bengio(2017)Warde-Farley and
  Bengio]{warde2016improving}
David Warde-Farley and Yoshua Bengio.
\newblock Improving generative adversarial networks with denoising feature
  matching.
\newblock In \emph{ICLR}, 2017.

\bibitem[Xiang \& Li(2017)Xiang and Li]{xiang2017effect}
Sitao Xiang and Hao Li.
\newblock On the effect of batch normalization and weight normalization in
  generative adversarial networks.
\newblock \emph{arXiv preprint arXiv:1704.03971}, 2017.

\bibitem[Yang et~al.(2017)Yang, Kannan, Batra, and Parikh]{yang2017lr}
Jianwei Yang, Anitha Kannan, Dhruv Batra, and Devi Parikh.
\newblock \rm{LR-GAN}: Layered recursive generative adversarial networks for
  image generation.
\newblock \emph{ICLR}, 2017.

\bibitem[Yoshida \& Miyato(2017)Yoshida and Miyato]{yoshida2017spectral}
Yuichi Yoshida and Takeru Miyato.
\newblock Spectral norm regularization for improving the generalizability of
  deep learning.
\newblock \emph{arXiv preprint arXiv:1705.10941}, 2017.

\end{thebibliography}

\clearpage

%%% Images here.

\clearpage
\appendix
\section{\label{apdsec:algo_snorm} The Algorithm of Spectral Normalization}
Let us describe the shortcut in Section~\ref{subsec:SN} in more detail.
We begin with vectors $\tilde{\bmu}$ that is randomly initialized for each weight.
If there is no multiplicity in the dominant singular values and if $\tilde{\bmu}$ is not orthogonal to the first left singular vectors\begin{footnote}{In practice, we are safe to assume that $\tilde{\bmu}$ generated from uniform distribution on the sphere is not orthogonal to the first singular vectors, because this can happen with probability $0$. }\end{footnote}, we can appeal to the principle of the power method and produce the first left and right singular vectors through the following update rule:
\begin{align}
	\tilde{\bmv} \leftarrow W^{\rm T} \tilde{\bmu}/\|W^{\rm T} \tilde{\bmu}\|_2,~
    \tilde{\bmu} \leftarrow W \tilde{\bmv}/\|W \tilde{\bmv}\|_2. \label{eq:poweriter}
\end{align}

We can then approximate the spectral norm of $W$ with the pair of so-approximated singular vectors:
\begin{align}
	\sigma(W) \approx  \tilde{\bmu}^{\rm T} W \tilde{\bmv}.
\end{align}

If we use SGD for updating $W$, the change in $W$ at each update would be small, and hence the change in its largest singular value.
In our implementation, we took advantage of this fact and reused the $\tilde{\bmu}$ computed at each step of the algorithm as the initial vector in the subsequent step.
In fact, with this `recycle' procedure, one round of power iteration was sufficient in the actual experiment to achieve satisfactory performance.
Algorithm~\ref{algo:snorm} in the appendix summarizes the computation of the spectrally normalized weight matrix $\bar{W}$ with this approximation.
Note that this procedure is very computationally cheap even in comparison to the calculation of the forward and backward propagations on neural networks.
Please see Figure~\ref{fig:cptime} for actual computational time with and without spectral normalization.

\begin{algorithm}
  \caption{\label{algo:snorm} SGD with spectral normalization}
  \begin{tight_itemize}
  \item Initialize $\tilde{\bmu}_l\in \mathcal{R}^{d_l}~{\rm for}~l=1,\dots,L$ with a random vector (sampled from isotropic distribution).
  \item For each update and each layer $l$:
  \begin{tight_enumerate}
  \item Apply power iteration method to a unnormalized weight $W^l$:
  \begin{align}
      \tilde{\bmv}_l &\leftarrow (W^{l})^{\rm T} \tilde{\bmu}_l/\|(W^{l})^{\rm T} \tilde{\bmu}_l\|_2\\
      \tilde{\bmu}_l &\leftarrow W^{l} \tilde{\bmv}_l/\|W^l \tilde{\bmv}_l\|_2
  \end{align}
  \item Calculate $\bar{W}_{\rm SN}$ with the spectral norm:
  \begin{align}
      \bar{W}_{\rm SN}^l(W^l) = W^l / \sigma(W^l),\ {\rm where}\ \sigma(W^l)=\tilde{\bmu}_l^{\rm T} W^l \tilde{\bmv}_l
  \end{align}
  \item Update $W^l$ with SGD on mini-batch dataset $\mathcal{D}_M$ with a learning rate $\alpha$:
  \begin{align}
      W^l \leftarrow W^l - \alpha \nabla_{W^l} \ell(\bar{W}_{\rm SN}^l(W^l), \mathcal{D}_M)
  \end{align}
  \end{tight_enumerate}
  \end{tight_itemize}
\end{algorithm}

\section{Experimental Settings}

\subsection{\label{apdsubsec:performance_measures}Performance measures}
Inception score is introduced originally by~\cite{salimans2016improved}: $I(\{x_n\}_{n=1}^N) := \exp(\E[D_{\rm KL}[p(y|\bmx)||p(y)]])$,
where $p(y)$ is approximated by $\frac{1}{N}\sum_{n=1}^N p(y|\bmx_n)$ and $p(y|x)$ is the trained Inception convolutional neural network \cite[]{szegedy2015going}, which we would refer to Inception model for short.
In their work,~\cite{salimans2016improved} reported that this score is strongly correlated with subjective human judgment of image quality.
Following the procedure in~\cite{salimans2016improved,warde2016improving}, we calculated the score for randomly generated 5000 examples from each trained generator to evaluate its ability to generate natural images.
We repeated each experiment 10 times and reported the average and the standard deviation of the inception scores.

Fr\'echet inception distance~\cite[]{heusel2017gans} is another measure for the quality of the generated examples that uses 2nd order information of the final layer of the inception model applied to the examples.
On its own, the \textit{Fre\'chet distance}~\cite{dowson1982frechet} is 2-Wasserstein distance between two distribution $p_1$ and $p_2$ assuming they are both multivariate Gaussian distributions:
\begin{align}
	F(p_1, p_2) = \|\bmmu_{p_1}- \bmmu_{p_2}\|_2^2 + {\rm trace}\left({C_{{p_1}} + C_{p_2} - 2(C_{p_1}C_{p_2})^{1/2}}\right),
\end{align}
where $\{\bmmu_{p_1}, C_{p_1}\}$, $\{\bmmu_{p_2}, C_{p_2}\}$ are the mean and covariance of samples from $q$ and $p$, respectively.
If $f_{\ominus} $ is the output of the final layer of the inception model before the softmax, the Fr\'echet inception distance (FID) between two distributions $p_1$ and $p_2$ on the images is the distance between $f_{\ominus}\circ p_1$ and  $f_{\ominus}\circ p_2$.
We computed the Fr\'echet inception distance between the true distribution and the generated distribution empirically over 10000 and 5000 samples.
Multiple repetition of the experiments did not exhibit any notable variations on this score.

\subsection{Image Generation on CIFAR-10 and STL-10}
For the comparative study, we experimented with the recent ResNet architecture of~\cite{gulrajani2017improved} as well as the standard CNN.
For this additional set of experiments, we used Adam again for the optimization and used the very hyper parameter used in~\cite{gulrajani2017improved} ($\alpha=0.0002, \beta_1=0, \beta_2=0.9,　n_{dis}=5$).
For our SN-GANs, we doubled the feature map in the generator from the original, because this modification achieved better results.
Note that when we doubled the dimension of the feature map for the WGAN-GP experiment, however, the performance deteriorated.

\subsection{\label{apdsubsec:exp_imagenet}Image Generation on ImageNet}
The images used in this set of experiments were resized to $128 \times 128$ pixels.
The details of the architecture are given in Table~\ref{tab:resnets_imagenet}.
For the generator network of conditional GANs, we used conditional batch normalization (CBN) \cite[]{dumoulin2016learned, de2017modulating}.
Namely we replaced the standard batch normalization layer with the CBN conditional to the label information $y \in \{1,\dots,1000\}$.
For the optimization, we used Adam with the same hyperparameters we used for ResNet on CIFAR-10 and STL-10 dataset.
We trained the networks with 450K generator updates, and applied linear decay for the learning rate after 400K iterations so that the rate would be 0 at the end.

\subsection{\label{subsec:networks} Network Architectures}
\begin{table}[ht]
	\caption{\label{tab:cnn_models} Standard CNN models for CIFAR-10 and STL-10 used in our experiments on image Generation.
    The slopes of all lReLU functions in the networks are set to $0.1$.}
   	\centering
    \begin{subtable}{.49\linewidth}
    	\centering
    	{\begin{tabular}{c}
			\toprule
			\midrule
		 	$z\in \bbR^{128} \sim \mathcal{N}(0, I)$ \\
           	\midrule
            dense $\rightarrow$ $M_g$ $\times$ $M_g$ $\times$ 512 \\
            \midrule
            4$\times$4, stride=2 deconv. BN 256 ReLU\\
			\midrule
            4$\times$4, stride=2 deconv. BN 128 ReLU\\
            \midrule
            4$\times$4, stride=2 deconv. BN 64 ReLU\\
            \midrule
            3$\times$3, stride=1 conv. 3 Tanh\\
            \midrule
			\bottomrule
		\end{tabular}}
        \caption{\label{tab:gen}Generator, $M_g=4$  for SVHN and CIFAR10, and $M_g=6$ for STL-10}
    \end{subtable}
    \begin{subtable}{.49\linewidth}
    	\centering
    	{\begin{tabular}{c}
			\toprule
			\midrule
		 	RGB image $x\in \bbR^{M\times M \times 3}$ \\
			\midrule
            3$\times$3, stride=1 conv 64 lReLU\\
            4$\times$4, stride=2 conv 64 lReLU\\
            \midrule
            3$\times$3, stride=1 conv 128 lReLU\\
            4$\times$4, stride=2 conv 128 lReLU\\
            \midrule
            3$\times$3, stride=1 conv 256 lReLU\\
            4$\times$4, stride=2 conv 256 lReLU\\
            \midrule
            3$\times$3, stride=1 conv. 512 lReLU\\
            \midrule
            dense $\rightarrow$ 1 \\
            \midrule
			\bottomrule
		\end{tabular}}
        \caption{\label{tab:dis_deep}Discriminator, $M=32$ for SVHN and CIFAR10, and $M=48$ for STL-10}
    \end{subtable}
\end{table}

\begin{figure}[ht]
	\begin{tabular}{cc}
    	\begin{minipage}{.2\textwidth}
            \centering
            \includegraphics[width=1.0\textwidth]{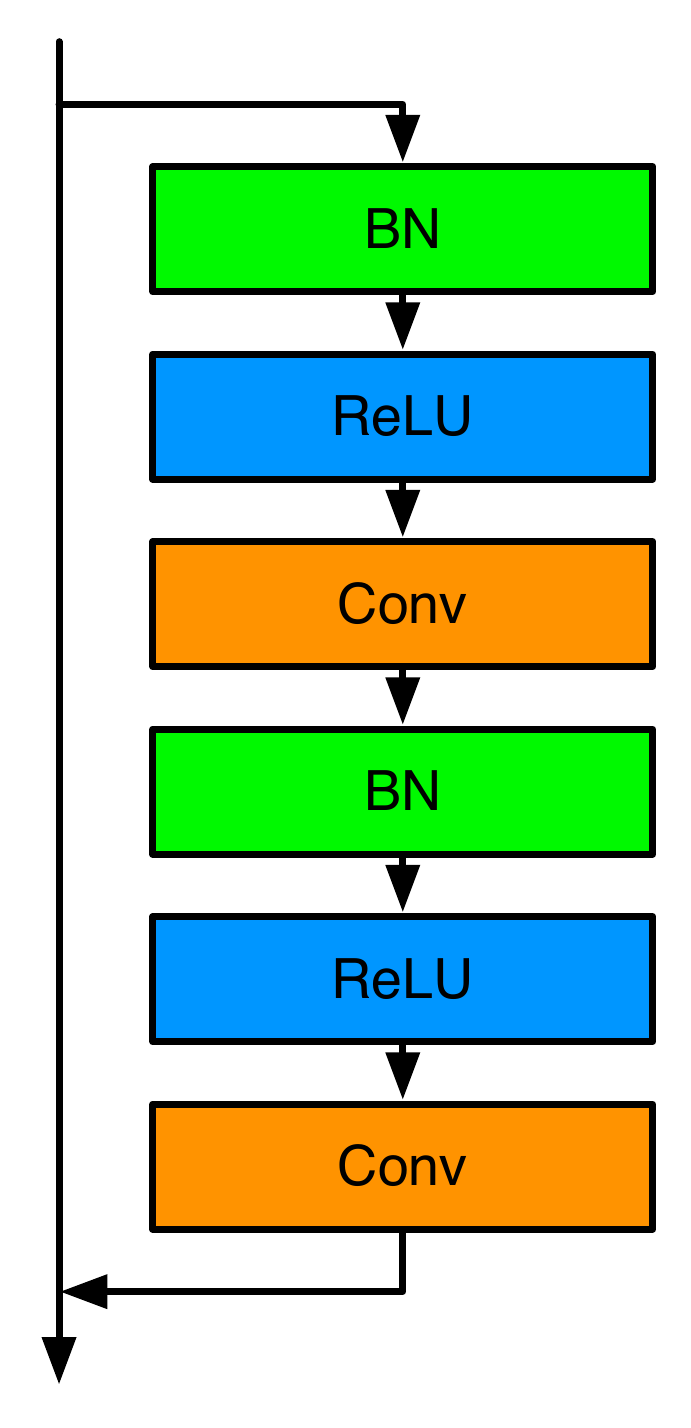}
            \figcaption{\label{fig:resblock} ResBlock architecture.
             For the discriminator we removed BN layers in ResBlock.}
        \end{minipage}
        \begin{minipage}{.75\textwidth}
          \tblcaption{\label{tab:resnets_cifar10}ResNet architectures for CIFAR10 dataset. We use similar architectures to the ones used in~\cite{gulrajani2017improved}. }
          \centering
          \begin{subtable}{.375\textwidth}
              \centering
              {\begin{tabular}{c}
                  \toprule
                  \midrule
                  $z\in \bbR^{128} \sim \mathcal{N}(0, I)$ \\
                  \midrule
                  dense, $4 \times 4 \times 256$ \\
                  \midrule
                  ResBlock up 256\\
                  \midrule
                  ResBlock up 256\\
                  \midrule
                  ResBlock up 256\\
                  \midrule
                  BN, ReLU, 3$\times$3 conv, 3 Tanh\\
                  \midrule
                  \bottomrule
              \end{tabular}}
              \caption{Generator}
          \end{subtable}\hspace{15mm}
          \begin{subtable}{.375\textwidth}
              \centering
              {\begin{tabular}{c}
                  \toprule
                  \midrule
                  RGB image $x\in \bbR^{32\times 32 \times 3}$ \\
                  \midrule
                  ResBlock down 128\\
                  \midrule
                  ResBlock down 128\\
                  \midrule
                  ResBlock 128\\
                  \midrule
                  ResBlock 128\\
                  \midrule
                  ReLU\\
                  \midrule
                  Global sum pooling\\
                  \midrule
                  dense $\rightarrow$ 1\\
                  \midrule
                  \bottomrule
              \end{tabular}}
              \caption{Discriminator}
          \end{subtable}
        \end{minipage}
    \end{tabular}
\end{figure}

\begin{table}[ht]
         \caption{\label{tab:resnets_stl}ResNet architectures for STL-10 dataset. }
          \centering
          \begin{subtable}{.375\textwidth}
              \centering
              {\begin{tabular}{c}
                  \toprule
                  \midrule
                  $z\in \bbR^{128} \sim \mathcal{N}(0, I)$ \\
                  \midrule
                  dense, $6 \times 6 \times 512$ \\
                  \midrule
                  ResBlock up 256\\
                  \midrule
                  ResBlock up 128\\
                  \midrule
                  ResBlock up 64\\
                  \midrule
                  BN, ReLU, 3$\times$3 conv, 3 Tanh\\
                  \midrule
                  \bottomrule
              \end{tabular}}
              \caption{Generator}
          \end{subtable}\hspace{15mm}
          \begin{subtable}{.375\textwidth}
              \centering
              {\begin{tabular}{c}
                  \toprule
                  \midrule
                  RGB image $x\in \bbR^{48\times 48 \times 3}$ \\
                  \midrule
                  ResBlock down 64\\
                  \midrule
                  ResBlock down 128\\
                  \midrule
                  ResBlock down 256\\
                  \midrule
                  ResBlock down 512\\
                  \midrule
                  ResBlock 1024\\
                  \midrule
                  ReLU\\
                  \midrule
                  Global sum pooling\\
                  \midrule
                  dense $\rightarrow$ 1 \\
                  \midrule
                  \bottomrule
              \end{tabular}}
              \caption{Discriminator}
          \end{subtable}
\end{table}

\begin{table}[ht]
         \caption{\label{tab:resnets_imagenet}ResNet architectures for image generation on ImageNet dataset.
         For the generator of conditional GANs, we replaced the usual batch normalization layer in the ResBlock with the conditional batch normalization layer. 
         As for the model of the \textit{projection discriminator}, we used the same architecture used in \cite{miyato2018cgans}.  Please see the paper for the details.}
          \centering
          \begin{subtable}{.3\textwidth}
              \centering
              {\begin{tabular}{c}
                  \toprule
                  \midrule
                  $z\in \bbR^{128} \sim \mathcal{N}(0, I)$ \\
                  \midrule
                  dense, $4 \times 4 \times 1024$ \\
                  \midrule
                  ResBlock up 1024\\
                  \midrule
                  ResBlock up 512\\
                  \midrule
                  ResBlock up 256\\
                  \midrule
                  ResBlock up 128\\
                  \midrule
                  ResBlock up 64\\
                  \midrule
                  BN, ReLU, 3$\times$3 conv 3 \\
                  \midrule
                  Tanh\\
                  \midrule
                  \bottomrule
              \end{tabular}}
              \caption{\label{tab:gen_resnet_imagenet} Generator}
          \end{subtable}
          \begin{subtable}{.3\textwidth}
              \centering
              {\begin{tabular}{c}
                  \toprule
                  \midrule
                  RGB image $x\in \bbR^{128\times 128 \times 3}$ \\
                  \midrule
                  ResBlock down 64\\
                  \midrule
                  ResBlock down 128\\
                  \midrule
                  ResBlock down 256\\
                  \midrule
                  ResBlock down 512\\
                  \midrule
                  ResBlock down 1024\\
                  \midrule
                  ResBlock 1024\\
                  \midrule
                  ReLU\\
                  \midrule
                  Global sum pooling\\
                  \midrule
                  dense $\rightarrow$ 1 \\
                  \midrule
                  \bottomrule
              \end{tabular}}
              \caption{\label{tab:dis_resnet_imagenet_unsup} Discriminator for unconditional GANs.}
          \end{subtable}\hspace{5mm}
          \begin{subtable}{.3\textwidth}
              \centering
              {\begin{tabular}{c}
                  \toprule
                  \midrule
                  RGB image $x\in \bbR^{128\times 128 \times 3}$ \\
                  \midrule
                  ResBlock down 64\\
                  \midrule
                  ResBlock down 128\\
                  \midrule
                  ResBlock down 256\\
                  \midrule
                  Concat(Embed($y$), $\bmh$)\\
                  \midrule
                  ResBlock down 512\\
                  \midrule
                  ResBlock down 1024\\
                  \midrule
                  ResBlock 1024\\
                  \midrule
                  ReLU\\
                  \midrule
                  Global sum pooling\\
                  \midrule
                  dense $\rightarrow$ 1 \\
                  \midrule
                  \bottomrule
              \end{tabular}}
              \caption{\label{tab:dis_resnet_imagenet} Discriminator for conditional GANs.
              For computational ease, we embedded the integer label $y\in\{0, \dots,1000\}$ into 128 dimension before concatenating the vector to the output of the intermediate layer.}
          \end{subtable}
\end{table}

\clearpage
\section{Appendix Results}
\subsection{\label{apdsubsec:acc_sn}Accuracy of spectral normalization}
\begin{figure}[ht]
  \centering
  \includegraphics[width=0.5\textwidth]{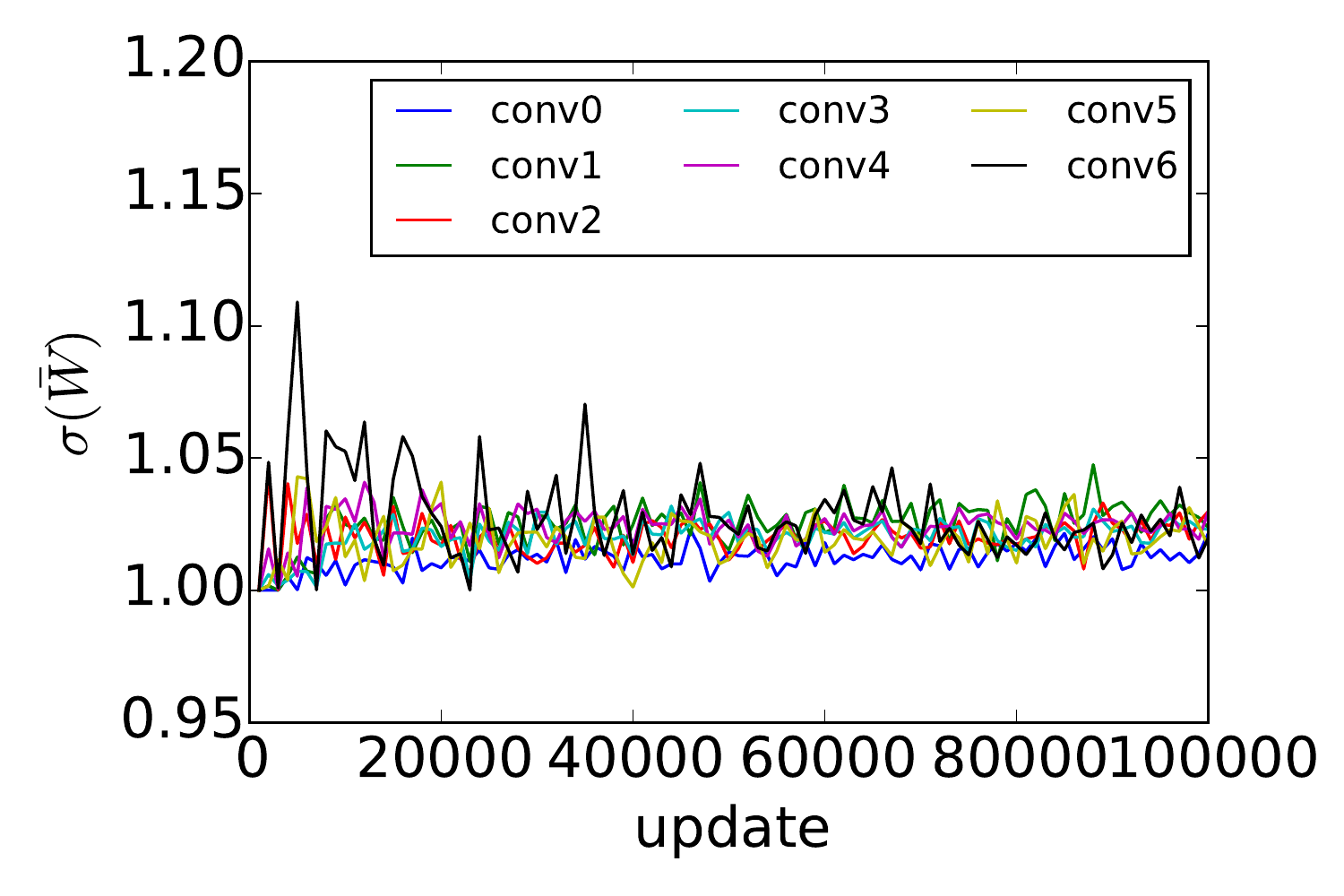}
  \figcaption{\label{fig:snorm} Spectral norms of all seven convolutional layers in the standard CNN during course of the training on CIFAR 10.}
\end{figure}

Figure~\ref{fig:snorm} shows the spectral norm of each layer in the discriminator over the course of the training.
The setting of the optimizer is C in Table~\ref{tab:settings} throughout the training.
In fact, they do not deviate by more than 0.05 for the most part. 
As an exception, 6 and 7-th convolutional layers with largest rank deviate by more than 0.1 in the beginning of the training, but the norm of this layer too stabilizes around $1$ after some iterations.

\subsection{Training Time}
\begin{figure}[ht]
\begin{tabular}{c}
\begin{minipage}{1.\textwidth}
	\centering
    \begin{subfigure}{0.3\textwidth}
		\includegraphics[width=1.0\textwidth]{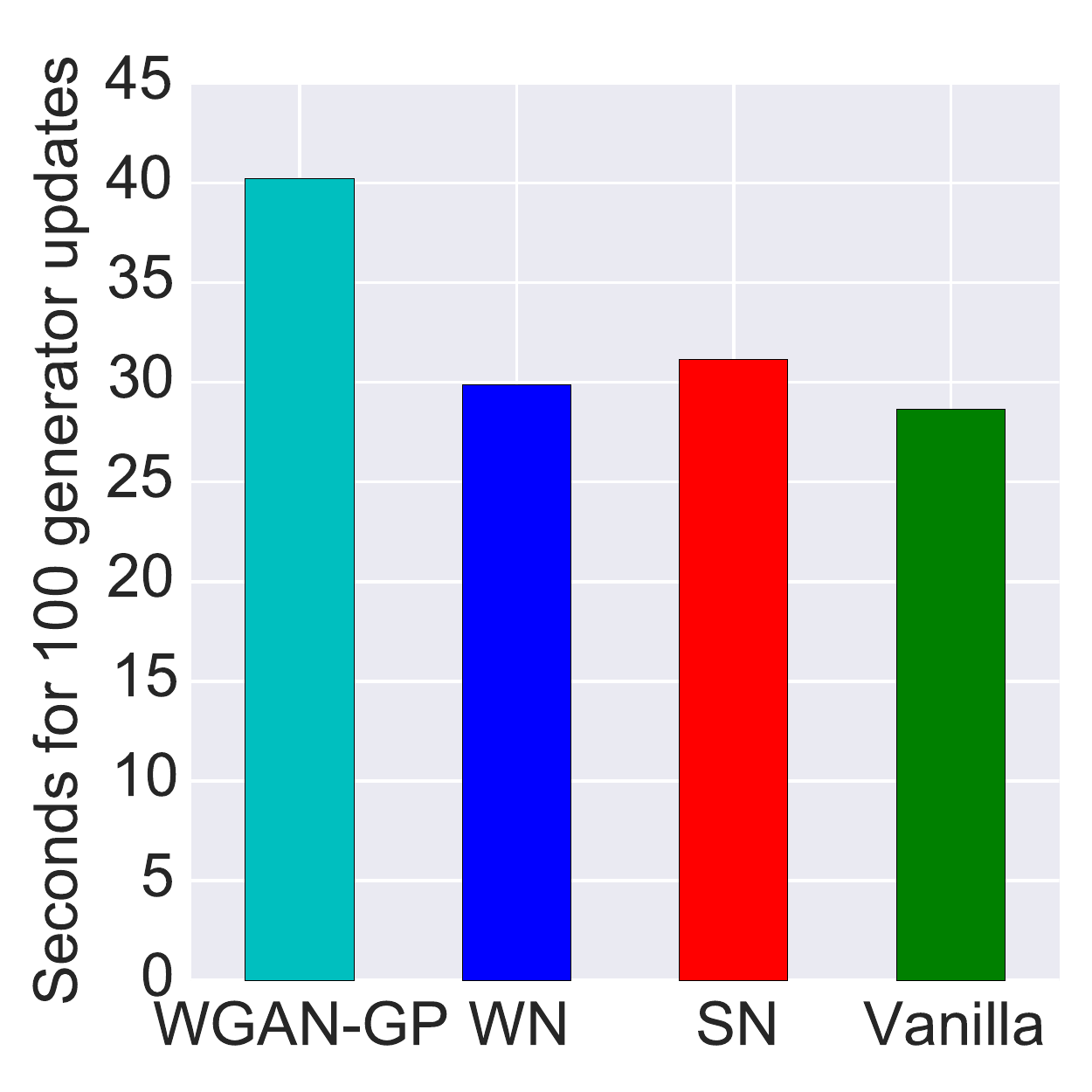}
        \caption{\label{fig:cptime_cnn_cifar10} CIFAR-10 (image size:$32\times32\times 3$)}
    \end{subfigure}
    \begin{subfigure}{0.3\textwidth}
		\includegraphics[width=1.0\textwidth]{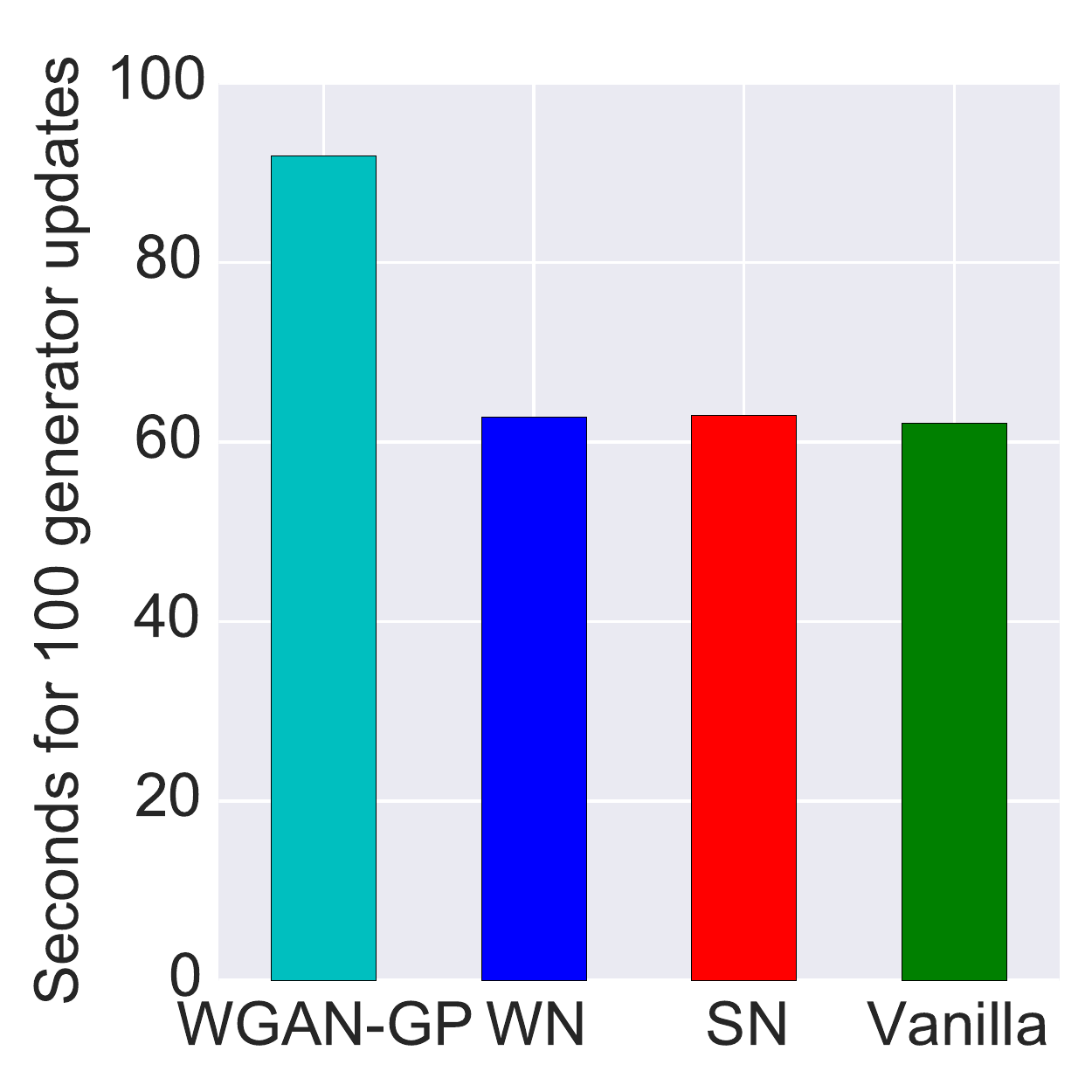}
        \caption{\label{fig:cptime_cnn_stl} STL-10 (images size:$48\times48\times 3$)}
    \end{subfigure}
    \figcaption{\label{fig:cptime}Computational time for 100 updates. We set $n_{\rm dis}=5$}
\end{minipage}
\end{tabular}
\end{figure}

\subsection{The effect of $n_{dis}$ on spectral normalization and weight normalization}
\begin{figure}[]
	\centering
	\includegraphics[width=.5\textwidth]{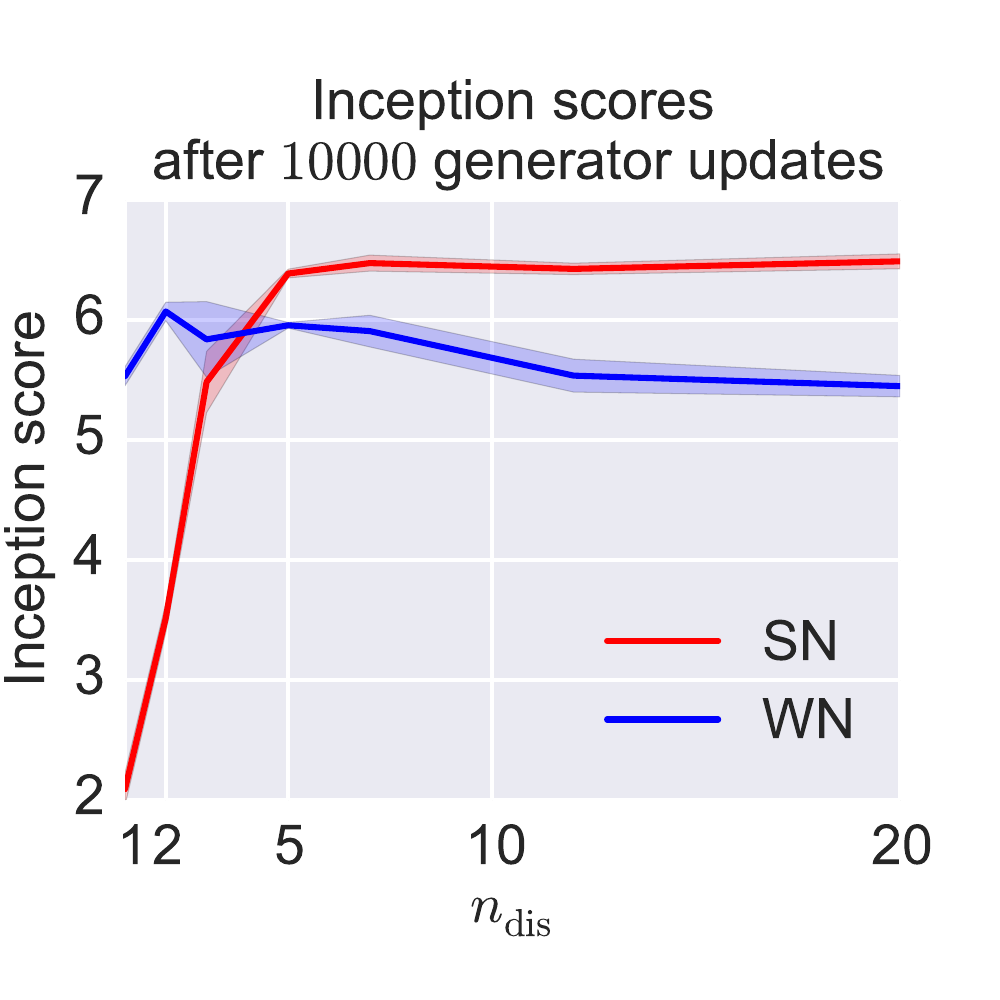}
    \figcaption{\label{fig:ndis_cifar10} The effect of $n_{dis}$ on spectral normalization and weight normalization. The shaded region represents the variance of the result over different seeds.}
\end{figure}

Figure~\ref{fig:ndis_cifar10} shows the effect of $n_{dis}$ on the performance of weight normalization and spectral normalization. All results shown in Figure~\ref{fig:ndis_cifar10} follows setting D, except for the value of $n_{dis}$.
For WN,
the performance deteriorates with larger $n_{dis}$, which amounts to computing minimax with better accuracy. Our SN does not suffer from this unintended effect.
% Analysis figures

\subsection{Generated Images on CIFAR10 with GAN-GP, Layer Normalization and Batch Normalization}

\begin{figure}[H]
	\centering
 	\includegraphics[width=1.0\textwidth]{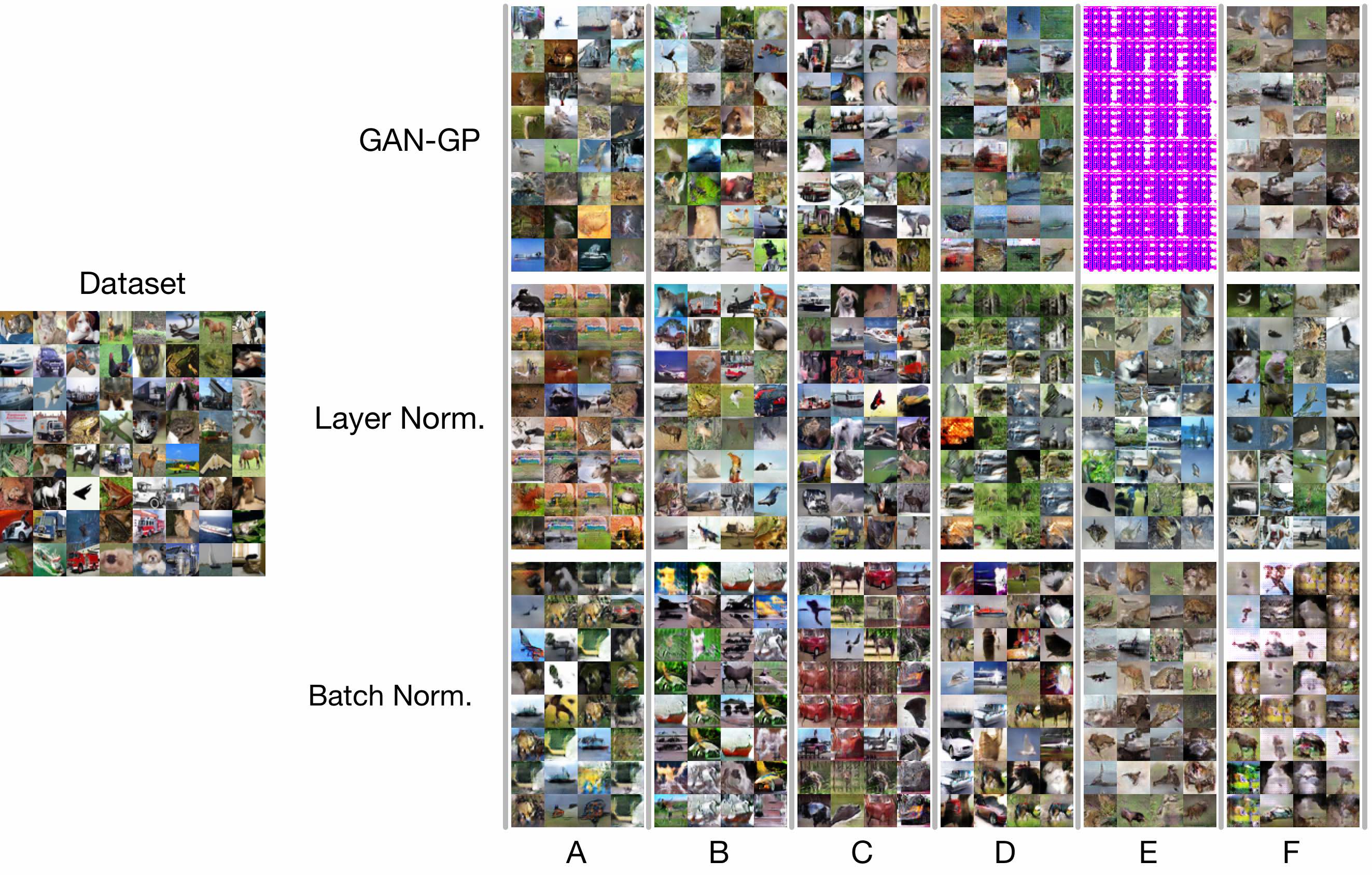}
    \caption{\label{fig:cifar10images_others} Generated images with GAN-GP, Layer Norm and Batch Norm on CIFAR-10}
\end{figure}

\clearpage

\subsection{Image generation on Imagenet}\label{apdsec:imagenet}
\begin{figure}[h]
	 \centering
     \begin{subfigure}{0.49\textwidth}
     	\centering
     	\includegraphics[width=0.7\textwidth]{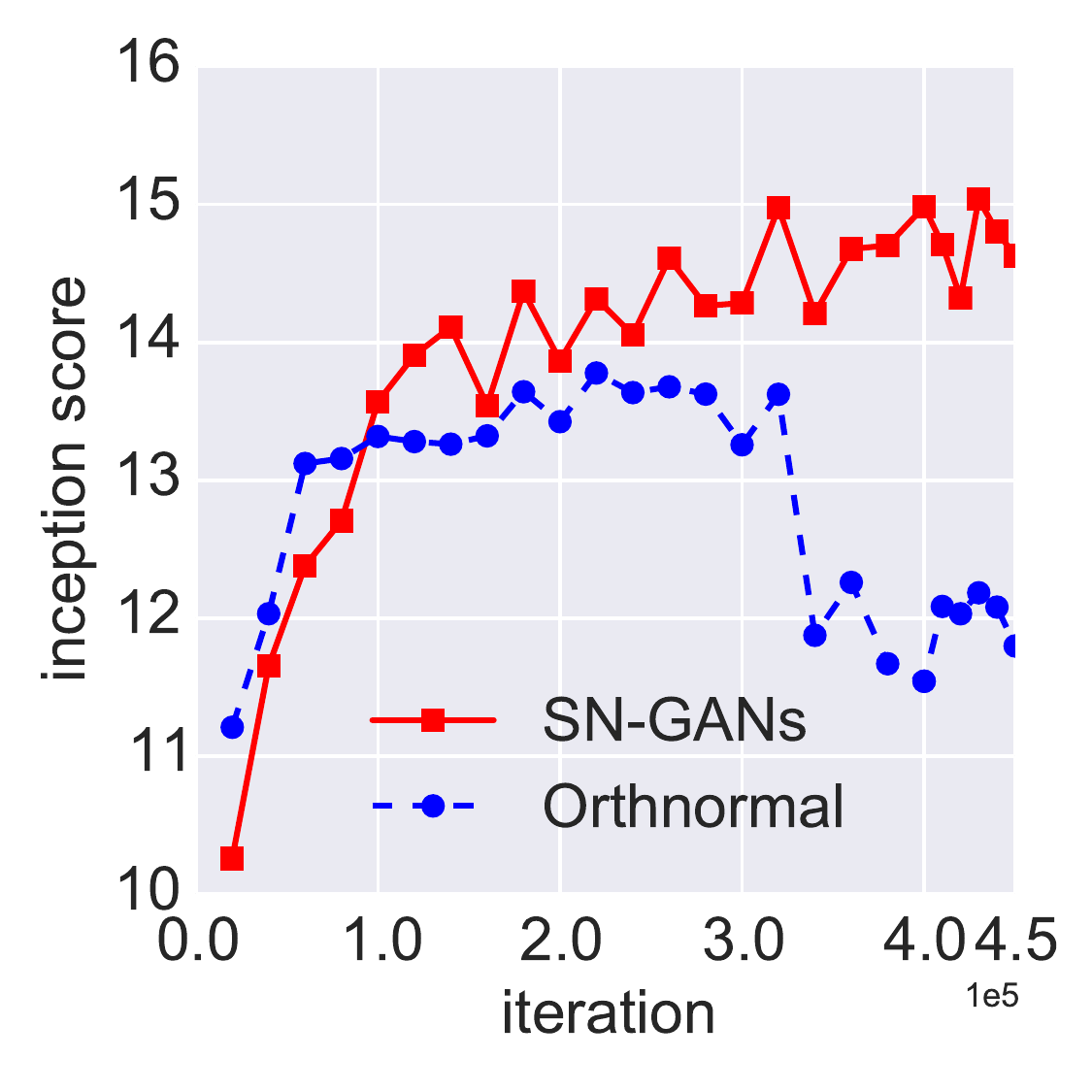}
        \figcaption{\label{fig:imagenet_incp_gan} Unconditional GANs}
     \end{subfigure}
     \begin{subfigure}{0.49\textwidth}
     	\centering
     	\includegraphics[width=0.7\textwidth]{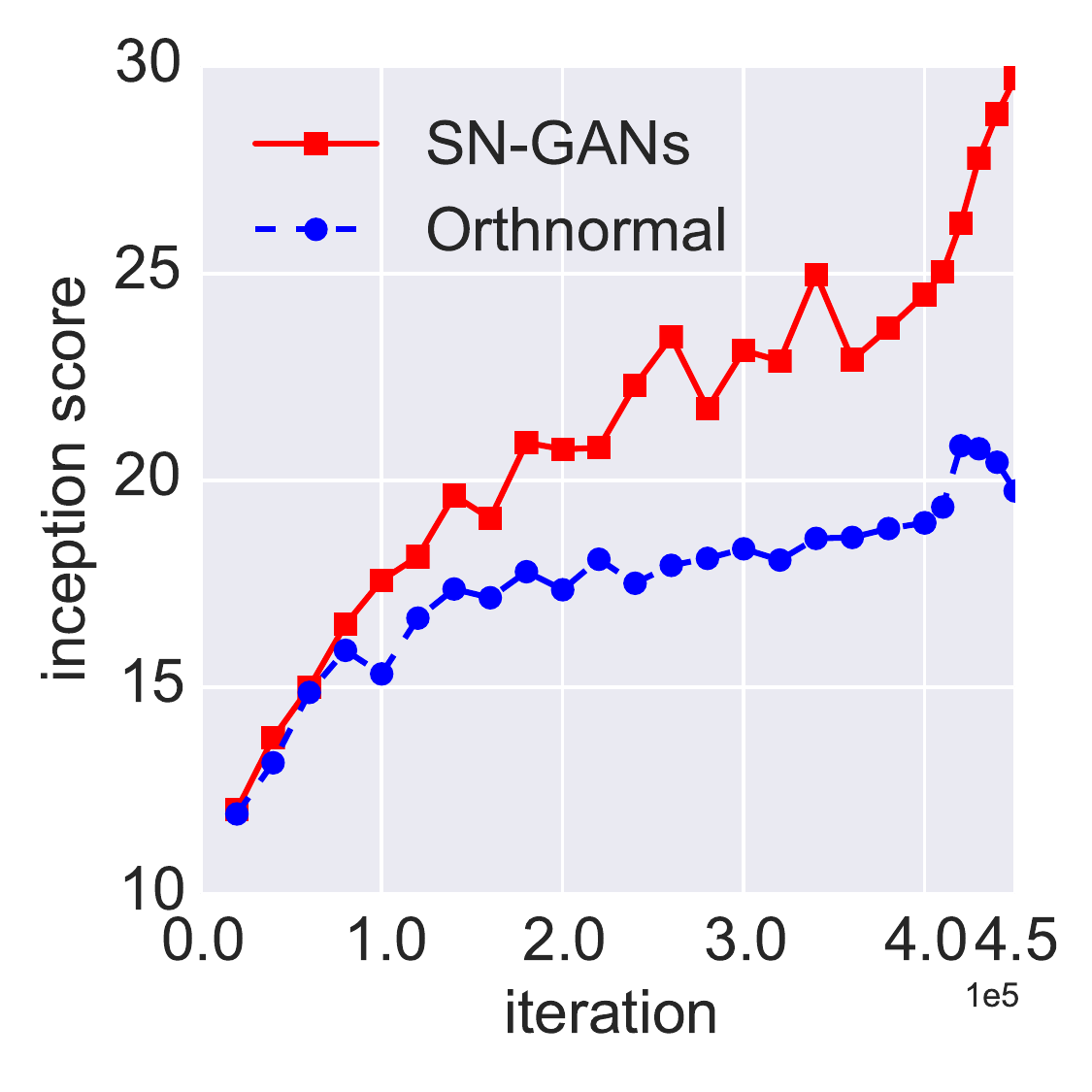}
        \figcaption{\label{fig:imagenet_incp_cgan_proj}Conditional GANs with \textit{projection discriminator}}
     \end{subfigure}
     \figcaption{\label{fig:imagenet_incp_others} Learning curves in terms of Inception score for SN-GANs and GANs with \textit{orthonormal regularization} on ImageNet. 
     The figure~(a) shows the results for the standard (unconditional) GANs, and the figure~(b) shows the results for the conditional GANs trained with ~\textit{projection discriminator}~\cite[]{miyato2018cgans}}
\end{figure}

\section{Spectral Normalization vs Other Regularization Techniques}
This section is dedicated to the comparative study of spectral normalization and other regularization methods for discriminators.
In particular, we will show that contemporary regularizations including weight normalization and weight clipping implicitly impose constraints on weight matrices that places unnecessary restriction on the search space of the discriminator.
More specifically, we will show that weight normalization and weight clipping \textit{unwittingly} favor low-rank weight matrices.
This can force the trained discriminator to be largely dependent on select few features, rendering the algorithm to be able to match the model distribution with the target distribution only on very low dimensional feature space.

\subsection{Weight Normalization and Frobenius Normalization}\label{subsubsec:wn_fn}
The weight normalization introduced by~\cite{salimans2016weight}
is a method that normalizes the $\ell_2$ norm of each row vector in the weight matrix\footnote{In the original literature, the weight normalization was introduced as a method for reparametrization of the form $\bar{W}_{\rm WN} := \left[\gamma_1\bar{\bmw}_1^{\rm T}, \gamma_2\bar{\bmw}_2^{\rm T},...,\gamma_{d_o}\bar{\bmw}_{d_o}^{\rm T}\right]^{\rm T}$ where $\gamma_i \in \bbR$ is to be learned in the course of the training.
In this work,  we deal with the case $\gamma_i=1 $ so that we can assess the methods under the Lipschitz constraint. }:

\begin{eqnarray}
	\bar{W}_{\rm WN} := \left[\bar{\bmw}_1^{\rm T}, \bar{\bmw}_2^{\rm T},...,\bar{\bmw}_{d_o}^{\rm T}\right]^{\rm T},~\text{where}~\bar{\bmw}_i(\bmw_i) := \bmw_i / \|\bmw_i\|_2, \label{eq:wn}
\end{eqnarray}
where $\bar{\bmw}_i$ and $\bmw_i$ are the $i$th row vector of  $\bar{W}_{\rm WN}$ and $W$, respectively.

Still another technique to regularize the weight matrix is to use the Frobenius norm:
\begin{align}
	\bar{W}_{\rm FN} := W / \|W\|_{F},
\end{align}
where $\|W\|_{F}:= \sqrt{\mathrm{tr}(W^{\rm T}W)} = \sqrt{\sum_{i,j} w_{ij}^2}$.

Originally, these regularization techniques were invented with the goal of improving the generalization performance of supervised training~\cite[]{salimans2016weight, arpit2016normalization}.
However, recent works in the field of GANs~\cite[]{salimans2016improved, xiang2017effect} found their another raison d'etat as a regularizer of discriminators, and succeeded in improving the performance of the original.

These methods in fact can render the trained discriminator $D$ to be $K$-Lipschitz for a some prescribed $K$ and achieve the desired effect to a certain extent.
However, weight normalization~\eqref{eq:wn} imposes the following implicit restriction on the choice of $\bar{W}_{\rm WN}$:
\begin{align}
	\sigma_1(\bar{W}_{\rm WN})^2 + \sigma_2(\bar{W}_{\rm WN})^2 + \cdots + \sigma_T(\bar{W}_{\rm WN})^2 = d_o,~{\rm where}~T=\min(d_i,d_o), \label{eq:svconstraint_apd}
\end{align}
where $\sigma_t(A)$ is a $t$-th singular value of matrix $A$.
The above equation holds because $\sum_{t=1}^{\min(d_i,d_o)} \sigma_t(\bar{W}_{\rm WN})^2 = \mathrm{tr}(\bar{W}_{\rm WN} \bar{W}_{\rm WN}^{\rm T})=\sum_{i=1}^{d_o} \frac{\bmw_i}{\|\bmw_i\|_2}\frac{\bmw_i^{\rm T}}{\|\bmw_i\|_2} = d_o$.
Under this restriction, the norm $\|\bar{W}_{\rm WN}\bmh\|_2$ for a fixed unit vector $\bmh$ is maximized at $\| \bar{W}_{\rm WN} \bmh \|_2 =\sqrt{d_o}$ when $\sigma_1(\bar{W}_{\rm WN})=\sqrt{d_o}$ and $\sigma_t(\bar{W}_{\rm WN}) = 0$ for $t=2,\dots,T$, which means that  $\bar{W}_{\rm WN}$ is of rank one.
Using such $W$ corresponds to using only one feature to discriminate the model probability distribution from the target.
Similarly, Frobenius normalization requires $\sigma_1(\bar{W}_{\rm FN})^2 + \sigma_2(\bar{W}_{\rm FN})^2 +\cdots +\sigma_T(\bar{W}_{\rm FN})^2 = 1$, and the same argument as above follows.

Here, we see a critical problem in these two regularization methods.
In order to retain as much norm of the input as possible and hence to make the discriminator more sensitive, one would hope to make the norm of $\bar{W}_{\rm WN} \bmh$ large.
For weight normalization,  however, this comes at the cost of reducing the rank and hence the number of features to be used for the discriminator.
Thus, there is a conflict of interests between weight normalization and our desire to use as many features as possible to distinguish the generator distribution from the target distribution.
The former interest often reigns over the other in many cases, inadvertently diminishing the number of features to be used by the discriminators.
Consequently, the algorithm would produce a rather arbitrary model distribution that matches the target distribution only at select few features.

Our spectral normalization, on the other hand, do not suffer from such a conflict in interest.
Note that the Lipschitz constant of a linear operator is determined only by the maximum singular value.
In other words, the spectral norm is independent of rank.
Thus, unlike the weight normalization, our spectral normalization
allows the parameter matrix to use as many features as possible while satisfying local $1$-Lipschitz constraint.
Our spectral normalization leaves more freedom in choosing the number of singular components (features) to feed to the next layer of the discriminator.

\begin{figure*}[ht]
	\centering
	\includegraphics[width=0.5\textwidth]{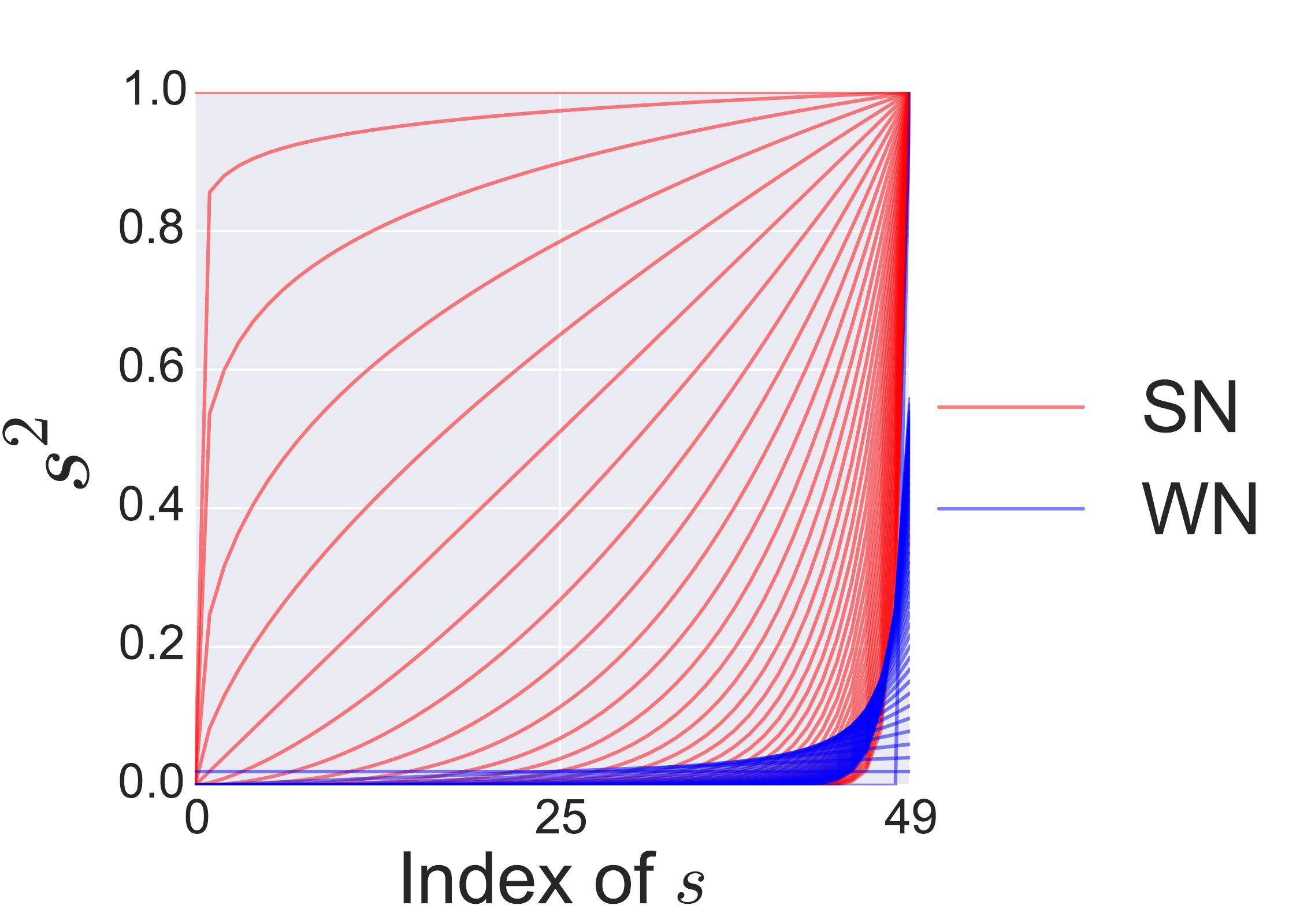}
    \caption{\label{fig:wn_vs_sn}
    \textbf{Visualization of the difference between spectral normalization (Red) and weight normalization (Blue) on possible sets of singular values.}
    The possible sets of singular values plotted in increasing order for weight normalization (Blue) and for spectral normalization (Red).
    For the set of singular values permitted under the spectral normalization condition, we scaled $\bar{W}_{\rm WN}$ by $1/\sqrt{d_o}$ so that its spectral norm is exactly 1.
    By the definition of the weight normalization, \textit{the area under the blue curves are all bound to be $1$}.
    Note that the range of choice for the weight normalization is small.}
\end{figure*}

%%%%%%%%%%%%%%%%%%%%%%%%%%%%%%%%%%
%%%%%%%%%%%%%%%%%%%%%%%%%%%%%%%%%%
To see this more visually, we refer the reader to Figure~\eqref{fig:wn_vs_sn}.
Note that spectral normalization allows for a wider range of choices than weight normalization.

In summary, weight normalization and Frobenius normalization favor skewed distributions of singular values, making the column spaces of the  weight matrices lie in (approximately) low dimensional vector spaces.
On the other hand, our spectral normalization does not compromise the number of feature dimensions used by the discriminator.
In fact, we will experimentally show that GANs trained with our spectral normalization can generate a synthetic dataset with wider variety and higher inception score than the GANs trained with other two regularization methods.

\subsection{Weight Clipping}
Still another regularization technique is \textit{weight clipping}
introduced by~\cite{arjovsky2017wasserstein} in their training of Wasserstein GANs.
Weight clipping simply truncates each element of weight matrices so that its absolute value is bounded above by a prescribed constant $c \in \bbR_+$.
Unfortunately, weight clipping suffers from the same problem as weight normalization and Frobenius normalization.
With weight clipping with the truncation value $c$, the value $\|W\bmx\|_2$ for a fixed unit vector $\bmx$ is maximized when the rank of $W$ is again one, and the training will again favor the discriminators that use only select few features. % to a very low dimensional function space.
\cite{gulrajani2017improved} refers to this problem as capacity underuse problem.
They also reported that the training of WGAN with weight clipping is slower than that of the original DCGAN~\cite[]{radford2015unsupervised}.
%~\cite{arjovsky2017wasserstein} では著者らは \textit{weight clipping} を用いてdiscrimininatorにLipschitz制約をもたせようとした。
%weight clipping method では、彼らは$D$の重みを毎回の更新のあとに重みの各成分を$[-c, c]$のレンジにおさまるようにクリップしている。
%しかし、weight clippingもweight normalizationやFRobenius normalizationと同様問題が存在する：clipped weightのspectral ノルムを最大になるときweightはランク１で、この線形変換にLipschitz制約を課したもとではweightが非常に制限される。
%この　capacity underuse problem on weight clipping は \cite{gulrajani2017improved}でも指摘されており、 $D$にweight clippingを用いた場合にはoriginal DCGAN~\cite{radford2015unsupervised}に比べて収束が 遅いと報告されている。

%\subsubsection{特異値クリッピング、もしくは特異値制約}

\subsection{Singular Value Clipping and Singular Value Constraint}
One direct and straightforward way of controlling the spectral norm is to clip the singular values~\cite[]{saito2016temporal}, ~\cite[]{jia2016improving}.
This approach, however, is computationally heavy because one needs to implement singular value decomposition in order to compute all the singular values.

A similar but less obvious approach is to parametrize $W \in \bbR^{d_o \times d_i}$ as follows from the get-go and train the discriminators with this constrained parametrization:
\begin{align}
	W := USV^{\rm T}, ~{\rm\ subject\ to\ } U^{\rm T} U = I,  V^{\rm T} V = I,~\max_{i} {S_{ii}} = K,
\end{align}
where $U\in \bbR^{d_{o} \times P}$, $V\in \bbR^{d_{i}\times P}$, and $S \in \bbR^{P \times P}$ is a diagonal matrix.
However, it is not a simple task to train this model while remaining absolutely faithful to this parametrization constraint.
Our spectral normalization, on the other hand, can carry out the updates with relatively low computational cost without compromising the normalization constraint.

\subsection{WGAN with Gradient Penalty (WGAN-GP)} \label{subsubsec:wgan-gp}
Recently,~\cite{gulrajani2017improved} introduced a technique to enhance the stability of the training of Wasserstein GANs~\cite[]{arjovsky2017wasserstein}.
In their work, they endeavored to place $K$-Lipschitz constraint~\eqref{eq:lipschitz} on the discriminator by augmenting the adversarial loss function with the following regularizer function:
\begin{align}
	\lambda \E_{\hat{\bmx} \sim p_{\hat{\bmx}}}[(\|\nabla_{\hat{\bmx}}D(\hat{\bmx})\|_2-1)^2], \label{eq:wgan_gp_reg}
\end{align}
where $\lambda > 0$ is a balancing coefficient and $\hat{\bmx}$ is:
\begin{align}
	\hat{\bmx} &:= \epsilon \bmx + (1-\epsilon) \tilde{\bmx} \\
    &{\rm where~}\epsilon \sim U[0,1],~\bmx \sim p_{\rm data}, ~\tilde{\bmx}=G(\bmz),~\bmz \sim p_\bmz.
\end{align}
Using this augmented objective function,~\cite{gulrajani2017improved} succeeded in training a GAN based on ResNet~\cite[]{he2016deep} with an impressive performance.
The advantage of their method in comparison to spectral normalization is that they can impose local $1$-Lipschitz constraint directly on the discriminator function without a rather round-about layer-wise normalization.
This suggest that their method is less likely to underuse the capacity of the network structure.

At the same time, this type of method that penalizes the gradients at sample points $\hat{\bmx}$ suffers from the obvious problem of not being able to regularize the function at the points outside of the support of the current generative distribution.
In fact, the generative distribution and its support gradually changes in the course of the training, and this can destabilize the effect of the regularization itself.

On the contrary, our spectral normalization regularizes the function itself, and the effect of the regularization is more stable with respect to the choice of the batch.
In fact, we observed in the experiment that a high learning rate can destabilize the performance of WGAN-GP. Training with our spectral normalization does not falter with aggressive learning rate.

Moreover, WGAN-GP requires more computational cost than our spectral normalization with single-step power iteration, because the computation of  $\|\nabla_{\bmx}D\|_2$ requires one whole round of forward and backward propagation.
In Figure~\ref{fig:cptime}, we compare the computational cost of the two methods for the same number of updates.

Having said that, one shall not rule out the possibility that the  gradient penalty can compliment spectral normalization and vice versa.　
Because these two methods regularizes discriminators by completely different means, and in the experiment section, we actually confirmed that combination of WGAN-GP and reparametrization with spectral normalization improves the quality of the generated examples over the baseline (WGAN-GP only).

\section{Reparametrization motivated by the spectral normalization}
\label{subsec:reparametrization}
We can take advantage of the regularization effect of the spectral normalization we saw above to develop another algorithm.
Let us consider another parametrization of the weight matrix of the discriminator given by:
\begin{important}{equation}
	\tilde{W}:= \gamma \bar{W}_{\rm SN} \label{eq:sn_reparam}
\end{important}
where $\gamma$ is a scalar variable to be learned.
This parametrization compromises the $1$-Lipschitz constraint at the layer of interest, but gives more freedom to the model while keeping the model from becoming degenerate.
For this reparametrization,  we need to control the Lipschitz condition by other means, such as the gradient penalty~\cite[]{gulrajani2017improved}.
Indeed, we can think of analogous versions of reparametrization by replacing $\bar{W}_{\rm SN}$ in \eqref{eq:sn_reparam} with $W$ normalized by other criterions.
The extension of this form is not new. In~\cite{salimans2016weight}, they originally introduced weight normalization in order to derive the reparametrization of the form~\eqref{eq:sn_reparam} with $\bar W_{\rm SN}$ replaced~\eqref{eq:sn_reparam} by $W_{\rm WN}$ and vectorized $\gamma$.

\subsection{Experiments: Comparison of Reparametrization with different normalization methods}

In this part of the addendum, we experimentally compare the reparametrizations derived from two different normalization methods (weight normalization and spectral normalization).
We tested the reprametrization methods for the training of the discriminator of WGAN-GP.
For the architecture of the network in WGAN-GP, we used the same CNN we used in the previous section.
For the ResNet-based CNN, we used the same architecture provided by~\cite[]{gulrajani2017improved} \footnote{We implement our method based on the open-sourced code provided by the author~\cite[]{gulrajani2017improved} \url{https://github.com/igul222/improved_wgan_training/blob/master/gan_cifar_resnet.py}}.

Tables~\ref{tab:wgan_gp_reparam1}, \ref{tab:wgan_gp_reparam2} summarize the result.
We see that our method significantly improves the inception score from the baseline on the regular CNN, and slightly improves the score on the ResNet based CNN.

\begin{figure}[h]
	\centering
    \begin{subfigure}{0.55\textwidth}
    \includegraphics[width=1.0\textwidth]{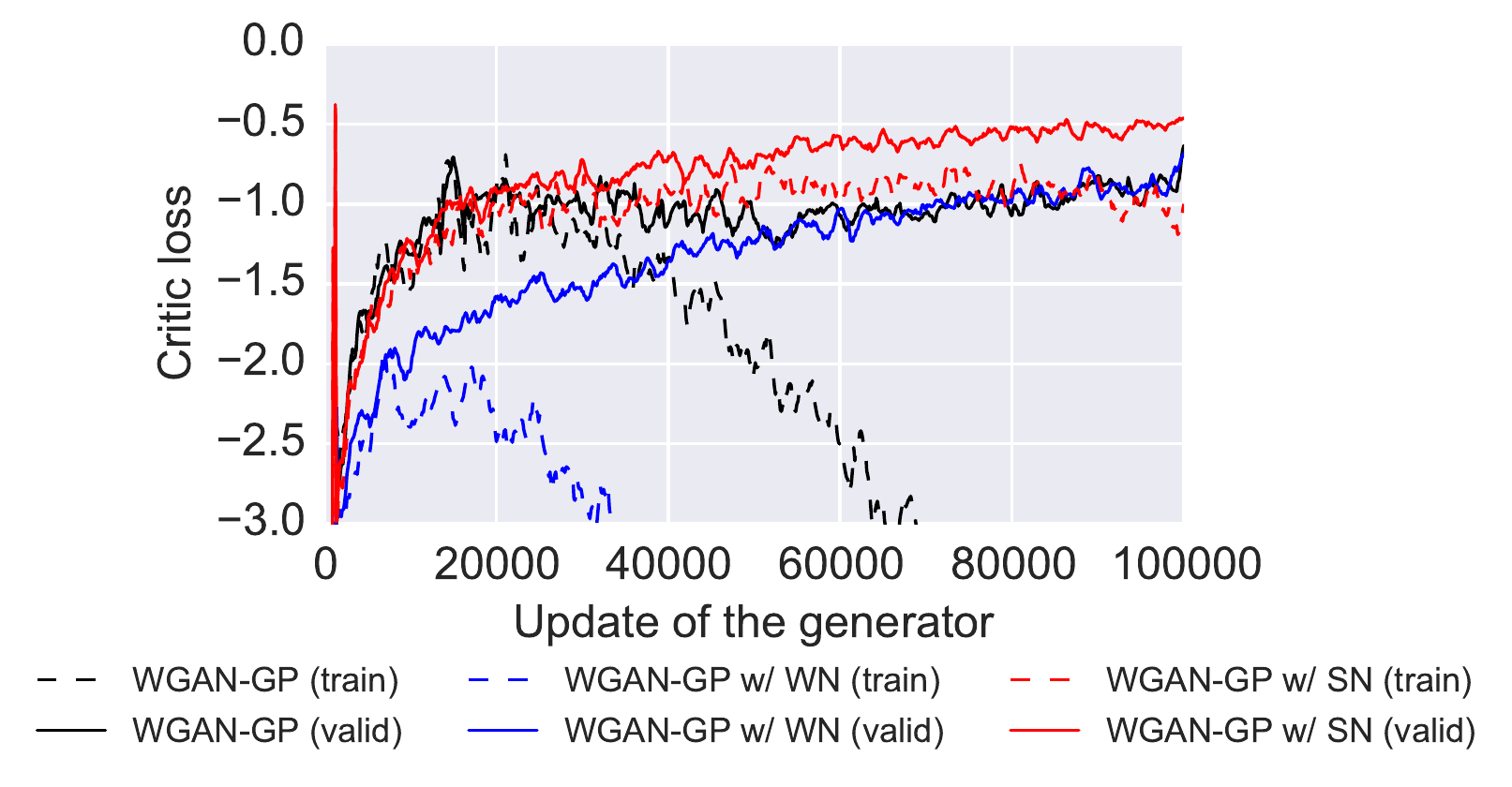}
    \caption{\label{subfig:critic_wgangp}Critic loss}
    \end{subfigure}
     \begin{subfigure}{0.4\textwidth}
    \includegraphics[width=1.0\textwidth]{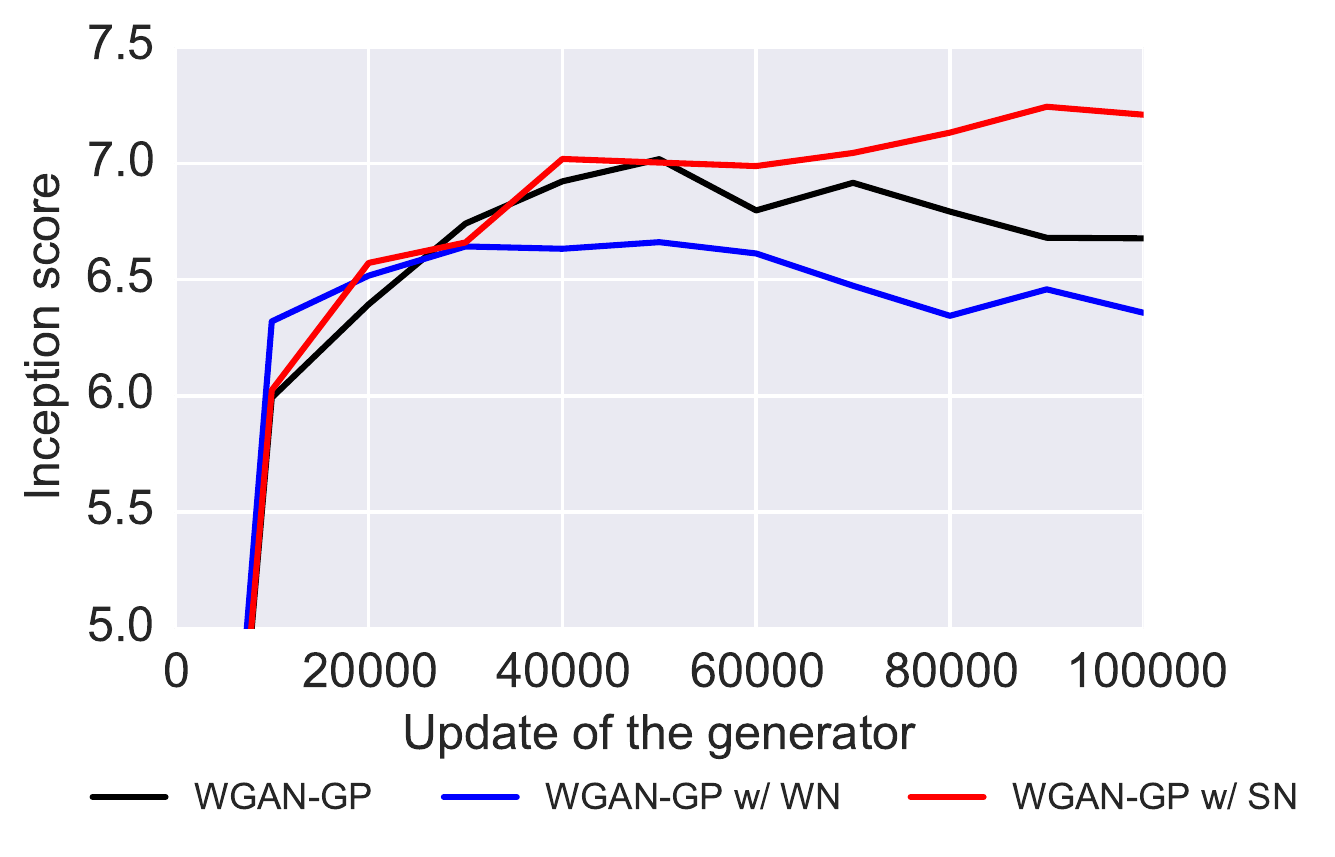}
    \caption{\label{subfig:incp_wgangp}Inception score}
    \end{subfigure}
    \caption{\label{fig:wgan_gp_reparam} Learning curves of (a) critic loss and (b) inception score on different reparametrization method on CIFAR-10 ; weight normalization (WGAN-GP w/ WN), spectral normalization (WGAN-GP w/ SN), and parametrization free (WGAN-GP).}
\end{figure}

Figure~\ref{fig:wgan_gp_reparam} shows the learning curves of (a) critic losses, on train and validation sets and (b) the inception scores with different reparametrization methods.
We can see the beneficial effect of spectral normalization in the learning curve of the discriminator as well.
We can verify in the figure ~\ref{subfig:critic_wgangp} that
the discriminator with spectral normalization overfits less to the training dataset than the discriminator without reparametrization and with weight normalization,
The effect of overfitting can be observed on inception score as well, and the final score with spectral normalization is better than the others.
As for the best inception score achieved in the course of the training, spectral normalization achieved 7.28, whereas the spectral normalization and vanilla normalization achieved 7.04 and 6.69, respectively.

\begin{table}
  \centering
  \begin{tabular}{lrr}
    \toprule
    \textbf{Method} & \textbf{Inception score} & \textbf{FID}\\
    \midrule
    WGAN-GP (Standard CNN, Baseline) & 6.68$\pm$.06 & 40.1  \\
    w/ Frobenius Norm. & N/A$^*$ & N/A$^*$ \\
    w/ Weight Norm. & 6.36$\pm$.04 & 42.4 \\
    w/ Spectral Norm. & \bf{7.20$\pm$.08} & \bf{32.0} \\
    \midrule
    (WGAN-GP, ResNet,~\cite{gulrajani2017improved}) & 7.86$\pm$.08\\
    WGAN-GP (ResNet, Baseline) & 7.80$\pm$.11 & 24.5 \\
    w/ Spectral norm. & 7.85$\pm$.06 & 23.6\\
    w/ Spectral norm. (1.5x feature maps in $D$) & \bf{7.96$\pm$.06} & \textbf{22.5} \\
    \bottomrule
  \end{tabular}
  \caption{\label{tab:wgan_gp_reparam1} Inception scores with different reparametrization mehtods on CIFAR10 without label supervisions. (*)We reported N/A for the inception score and FID of Frobenius normalization because the training collapsed at the early stage. }
\end{table}

\begin{table}
  \centering
  \begin{tabular}{lrr}
    \toprule
    \textbf{Method} (ResNet) & \textbf{Inception score} & \textbf{FID}\\
    \midrule
    (AC-WGAN-GP,~\cite{gulrajani2017improved}) & 8.42$\pm$.10 & \\
    \midrule
    AC-WGAN-GP (Baseline) & 8.29$\pm$.12 & 19.5 \\
    w/ Spectral norm. & 8.59$\pm$.12 & 18.6 \\
    w/ Spectral norm. (1.5x feature maps in $D$) & \textbf{8.60$\pm$.08} & \textbf{17.5} \\
    \bottomrule
  \end{tabular}
  \caption{\label{tab:wgan_gp_reparam2} Inception scores and FIDs with different reparametrization methods on CIFAR10 with the label supervision, by auxiliary classifier~\cite[]{odena2016conditional}. }
\end{table}

\section{The gradient of General Normalization Method}
Let us denote $\bar{W}:=W/N(W)$ to be the normalized weight where $N(W)$ to be a scalar normalized coefficient (e.g. Spectral norm or Frobenius norm).
In general, we can write the derivative of loss with respect to unnormalized weight $W$ as follows:
\begin{align}
	\frac{\partial V(G,D(W))}{\partial W} &= \frac{1}{N(W)}\left(\frac{\partial V}{\partial \bar{W}} - {\rm trace}\left( \left(\frac{\partial V}{\partial \bar{W}}\right)^{\rm T} \bar{W}\right) \frac{\partial (N(W))}{\partial W}\right) \\
     &=\frac{1}{N(W)} \left(\nabla_{\bar{W}}V - {\rm trace}\left((\nabla_{\bar{W}}V) ^{\rm T} \bar{W}\right) \nabla_W N\right) \\
     & = \alpha \left(\nabla_{\bar{W}}V - \lambda \nabla_W N\right),
\end{align}
where $\alpha:=1/N(W)$ and $\lambda:={\rm trace}\left((\nabla_{\bar{W}}V) ^{\rm T} \bar{W}\right)$.
The gradient $\nabla_{\bar{W}}V$ is calculated by $\hat{\E}\left[\bmdelta \bmh^{\rm T}\right]$ where  $\bmdelta := \left(\partial V(G,D) / \partial \left(\bar{W} \bmh\right)\right)^{\rm T}$, $\bmh$ is the hidden node in the network to be transformed by $\bar{W}$ and $\hat{\E}$ represents empirical expectation over the mini-batch.
When $N(W):=\|W\|_F$, the derivative is:
\begin{align}
  \frac{\partial V(G,D(W))}{\partial W} &= \frac{1}{\|W\|_F}\left(\hat{\E}\left[\bmdelta \bmh^{\rm T}\right] - {\rm trace}\left(\hat{\E}\left[\bmdelta \bmh^{\rm T}\right]^{\rm T} \bar{W}\right)\bar{W}\right),
\end{align}
and when $N(W):=\|W\|_2 = \sigma(W)$,
\begin{align}
  \frac{\partial V(G,D(W))}{\partial W} &= \frac{1}{\sigma(W)}\left(\hat{\E}\left[\bmdelta \bmh^{\rm T}\right] - {\rm trace}\left(\hat{\E}\left[\bmdelta \bmh^{\rm T}\right]^{\rm T} \bar{W}\right) \bmu_1 \bmv_1^{\rm T}\right).
\end{align}

Notice that, at least for the case $N(W):=\|W\|_{F}$ or $N(W):=\|W\|_{2}$, the point of this gradient is given by :
\begin{align}
	\nabla_{\bar{W}}V = k \nabla_{W}N.
\end{align}
where $^\exists k\in \bbR$

\end{document}